\documentclass[10pt,letterpaper]{article}
\usepackage[margin=1in]{geometry}

\usepackage{amssymb}
\usepackage{amsthm}

\usepackage{amsmath,amssymb,amsfonts}
\usepackage{graphicx}
\usepackage{textcomp}
\usepackage{xcolor}
\usepackage{contour}
\usepackage[margin=1in]{geometry}
\usepackage[T1]{fontenc}    
\usepackage{hyperref}       
\usepackage{url}            
\usepackage{booktabs}       
\usepackage{amsfonts}       
\usepackage{nicefrac}       
\usepackage{microtype}      
\usepackage[center]{subfigure} 
\usepackage{graphicx}  
\usepackage{amsmath}
\usepackage{xspace}
\usepackage{amssymb}
\usepackage{pifont}
\usepackage{stmaryrd} 

\usepackage{chngcntr}

\usepackage[skip=1pt]{caption}
\usepackage{enumitem}
\setitemize{noitemsep,topsep=0pt,parsep=0pt,partopsep=0pt}
\usepackage{tabularx}
\usepackage{rotating}
\usepackage{tablefootnote}
\usepackage{threeparttable}
\usepackage{xcolor}
\usepackage{wasysym}
\usepackage{multicol}
\usepackage{multirow}
\usepackage{mathtools}
\usepackage{esvect}
\usepackage{pdflscape}

\usepackage{algorithm}
\usepackage{algpseudocode}
\usepackage{makecell}

\usepackage{tikz}

\newtheorem{problem}{Problem}

\newtheorem{definition}{Definition}
\newtheorem{theorem}{Theorem}
\newtheorem*{proof*}{Proof}

\definecolor{airforceblue}{rgb}{0.36, 0.54, 0.66}
\definecolor{cobalt}{rgb}{0.0, 0.28, 0.67}
\definecolor{fgreen}{HTML}{228B22}

\newcommand{\smallsection}[1]{\noindent\underline{\smash{\textbf{#1:}}}}

\newcommand{\defeq}{\vcentcolon=}
\newcommand{\mat}[1]{\mathbf{#1}}

\newcommand{\vect}[1]{\mathbf{#1}}

\newcommand{\set}[1]{#1}

\newcommand{\PU}[1]{\mat{P}_{\users}^{(#1)}}
\newcommand{\PV}[1]{\mat{P}_{\items}^{(#1)}}
\newcommand{\MU}[1]{\mat{M}_{\users}^{(#1)}}
\newcommand{\MV}[1]{\mat{M}_{\items}^{(#1)}}
\newcommand{\XU}{\mat{X}_{\users}}
\newcommand{\XV}{\mat{X}_{\items}}

\newcommand{\hPU}[1]{\hat{\mat{P}}_{\users}^{(#1)}}
\newcommand{\hPV}[1]{\hat{\mat{P}}_{\items}^{(#1)}}
\newcommand{\hMU}[1]{\hat{\mat{M}}_{\users}^{(#1)}}
\newcommand{\hMV}[1]{\hat{\mat{M}}_{\items}^{(#1)}}

\newcommand{\graph}{\mathcal{G}}
\newcommand{\users}{\mathcal{U}}
\newcommand{\items}{\mathcal{V}}
\newcommand{\edges}{\mathcal{E}}

\newcommand{\lram}{\texttt{RMP}_{k}}

\newcommand{\matR}[1]{
    \mat{R}\raisebox{-0.3ex}{$\scriptstyle \users \overset{#1}{\rightarrow} \items$}
}
\newcommand{\mattR}[1]{
    \mat{\tilde{R}}\raisebox{-0.3ex}{$\scriptstyle \users \overset{#1}{\rightarrow} \items$}
}
\newcommand{\matRT}[1]{
    \mat{R}\raisebox{-0.3ex}{$\scriptstyle \items \overset{#1}{\leftarrow} \users$}
}
\newcommand{\mattRT}[1]{
    \mat{\tilde{R}}\raisebox{-0.3ex}{$\scriptstyle \items \overset{#1}{\leftarrow} \users$}
}

\newcommand{\matQ}[1]{
    \mat{Q}\raisebox{-0.3ex}{$\scriptstyle \items \overset{#1}{\rightarrow} \users$}
}
\newcommand{\mattQ}[1]{
    \mat{\tilde{Q}}\raisebox{-0.3ex}{$\scriptstyle \items \overset{#1}{\rightarrow} \users$}
}
\newcommand{\matQT}[1]{
    \mat{Q}\raisebox{-0.3ex}{$\scriptstyle \users \overset{#1}{\leftarrow} \items$}
}
\newcommand{\mattQT}[1]{
    \mat{\tilde{Q}}\raisebox{-0.3ex}{$\scriptstyle \users \overset{#1}{\leftarrow} \items$}
}

\newcommand{\matA}[1]{
    \mat{A}\raisebox{-0.3ex}{$\scriptstyle \users \overset{#1}{\rightarrow} \users$}
}
\newcommand{\mattA}[1]{
    \mat{\tilde{A}}\raisebox{-0.3ex}{$\scriptstyle \users \overset{#1}{\rightarrow} \users$}
}

\newcommand{\mattAT}[1]{
    \mat{\tilde{A}}\raisebox{-0.3ex}{$\scriptstyle \users \overset{#1}{\leftarrow} \users$}
}

\newcommand{\matabsR}{
    \mat{R}_{\users \rightarrow \items}
}
\newcommand{\matabsQ}{
    \mat{Q}_{\items \rightarrow \users}
}
\newcommand{\matabsA}{
    \mat{A}_{\users \rightarrow \users}
}

\let\oldsum\sum
\renewcommand{\sum}{\displaystyle\oldsum} 

\newcommand{\method}{\textsc{ELISE}\xspace}
\newcommand{\methods}{\textsc{ELISE-S}\xspace}
\newcommand{\methodr}{\textsc{ELISE-R}\xspace}

\begin{document}



\title{Effective and Lightweight Representation Learning for Link Sign Prediction in Signed Bipartite Graphs}

\author{Gyeongmin Gu\,$^{1,*}$, Minseo Jeon\,$^{2,*}$, Hyun-Je Song\,$^{1}$, Jinhong Jung\,$^{2,\dagger}$}
\date{\normalsize $^{1}$ School of Electronics and Information Engineering (Computer Science),\\ Jeonbuk National University, Jeonju, South Korea\\ 
	$^{2}$ School of Software, Soongsil University, Seoul, South Korea, \\
        \vspace{1mm}
	gyeongmin.gu@jbnu.ac.kr, minseojeon@soongsil.ac.kr, hyunje.song@jbnu.ac.kr, jinhong@ssu.ac.kr
}

\maketitle
\def\thefootnote{$*$}\footnotetext{Two first authors have contributed equally to this work.}
\def\thefootnote{$\dagger$}\footnotetext{Corresponding author.}
\def\thefootnote{\arabic{footnote}}






        

\begin{abstract}
How can we effectively and efficiently learn node representations in signed bipartite graphs?
A signed bipartite graph is a graph consisting of two nodes sets where nodes of different types are positively or negative connected, and it has been extensively used to model various real-world relationships such as e-commerce, peer review systems, etc.
To analyze such a graph, previous studies have focused on designing methods for learning node representations using graph neural networks (GNNs).
In particular, these methods insert edges between nodes of the same type based on balance theory, enabling them to leverage augmented structures in their learning.
However, the existing methods rely on a naive message passing design, which is prone to over-smoothing and susceptible to noisy interactions in real-world graphs.
Furthermore, they suffer from computational inefficiency due to their heavy design and the significant increase in the number of added edges.

In this paper, we propose \method, an effective and lightweight GNN-based approach for learning signed bipartite graphs.
We first extend personalized propagation to a signed bipartite graph, incorporating signed edges during message passing. 
This extension adheres to balance theory without introducing additional edges, mitigating the over-smoothing issue and enhancing representation power.
We then jointly learn node embeddings on a low-rank approximation of the signed bipartite graph, which reduces potential noise and emphasizes its global structure, further improving expressiveness without significant loss of efficiency.
We encapsulate these ideas into \method, designing it to be lightweight, unlike the previous methods that add too many edges and cause inefficiency.
Through extensive experiments on real-world signed bipartite graphs, we demonstrate that \method outperforms its competitors for predicting link signs while providing faster training and inference time.
\end{abstract}



\section{Introduction}
\label{sec:introduction}
How can we effectively and efficiently learn node representations in signed bipartite graphs?
A signed bipartite graph~\cite{DerrJCT19} is a graph with two disjoint node sets and edges between them, each edge having a positive or negative sign.
For example, in e-commerce, users and items are represented as nodes, with signed relationships showing whether a user likes or dislikes an item~\cite{DerrJCT19}.  
Thus, signed bipartite graphs can model real-world relationships between different types of nodes, and analyzing them is crucial for making useful predictions across various domains, such as e-commerce platforms~\cite{LeeKSJ24}, recommender systems~\cite{ParkJK17}, peer review systems, congressional voting, and social media.

Node representation learning~\cite{GroverL16} aims to embed nodes into continuous vector spaces by considering the structure of a given graph and plays an important role in analyzing the graph.
Among various models, graph neural networks (GNNs)~\cite{ScarselliGTHM09} have attracted considerable attention for learning representations in graphs.
In particular, various GNN-based methods~\cite{derrICDM2018,HuangSHC19} have been proposed for signed unipartite graphs, which consist of only one set of nodes.
To fully harness signed relationships, these methods exploit structural balance theory~\cite{cartwright1956structural}, aiming to ensure that embeddings of potential friendly nodes are similar, while those of potential hostile nodes are dissimilar.
However, the traditional balance theory analyzes signed triangles between nodes of the same type; thus, it is limited in explaining signed bipartite graphs, where no triangle is formed.

To analyze signed bipartite graphs, Derr et al.~\cite{DerrJCT19} extend the balance theory to consider quadruplet relationships, known as signed butterflies, between nodes of different types, and demonstrate the balance in signed bipartite graphs.
Using this, GNN-based methods~\cite{huangCIKM2021,zhangSIGIR2023} have recently been proposed to learn node representations in signed bipartite graphs.
Huang et al.~\cite{huangCIKM2021} claimed that a signed bipartite graph is extremely sparse as there are no edges between nodes of the same type. 
To leverage potential relationships, they inserted signed edges between nodes of the same type based on the balance theory and proposed SBGNN, which learns node embeddings by message passing between nodes of different types or within the same type.
Zhang et al.~\cite{zhangSIGIR2023} proposed SBGCL, which applies graph contrastive learning with augmentation techniques tailored to the signed graph constructed from both the original and new edges, similar to SBGNN.

However, the performance of these methods for learning signed bipartite graphs remains limited due to the following:
\begin{itemize}[noitemsep]
    \item {
    \smallsection{Naive message-passing design}
    The previous methods adopt the GNN architecture and naively rely on a simplistic  message-passing design. 
    However, this is vulnerable to the over-
    smoothing issue~\cite{LiHW18}, which makes embeddings indistinguishable as the number of layers increases due to the averaging operation in message-passing.
    Furthermore, real-world graphs contain noisy interactions (e.g., casually viewing items)~\cite{ouyang2016noise,shuCIKM2021}, which lead the message-passing process to aggregate erroneous neighbor information. 
    These issues prevent the existing methods from yielding more expressive representations.
    }
    \item {
    \smallsection{Computational inefficiency} 
    As described above, these methods insert a significant number of edges to consider the potential relationships between nodes of the same type. 
    Moreover, due to the simplistic adoption of the GNN architecture, they require weight matrices for various types of message-passing in each layer. 
    Furthermore, they optimize auxiliary loss functions~\cite{YouCSCWS20,liAAAI2020}, such as contrastive learning or balance regularization.
    Hence, these degrade efficiency and increases model complexity, requiring significant computational costs in terms of time and space.
    %
    }
\end{itemize}

In this paper, we propose \method (Effective and Lightweight Learning for Signed Bipartite Graphs), a new method for learning node representations in signed bipartite graphs. 
We propose \textit{signed personalized message passing} specialized for signed bipartite graphs, which injects personalized features during the message-passing step between nodes of different types while adhering to the balance theory.
This aims to make the resulting embeddings distinguishable while considering signed edges, thereby resolving the over-smoothing issue.
We further design \textit{refined message passing}, which performs message-passing on a low-rank approximation of the given graph, refining the graph and reducing the impact of noisy interactions.
We jointly learn the encoders based on these message-passing steps and design our model to be lightweight by avoiding the addition of edges and excluding additional weight matrices at each layer and auxiliary loss optimizations.

Our main contributions are summarized as follows:
\begin{itemize}
    \item {
        \textbf{Effective encoders.}
        We propose two encoders based on signed personalized and refined message passing. 
        The first encoder aims to alleviate the over-smoothing issue by injecting personalized features during its message passing over layers. 
        The second encoder aims to reduce the impact of noisy interactions by performing message passing on a low-rank approximation of a given signed bipartite graph.
        %
        
    }
    \item {
        \textbf{Lightweight encoder design.}
        We encapsulate our encoders into \method, designing it to be lightweight by avoiding the addition of edges between nodes of the same type and excluding additional weight matrices at each layer, as well as auxiliary loss optimizations, unlike the previous methods for signed bipartite graphs.
    }
    \item {
        \textbf{Experiments.}
        We conduct extensive experiments on four real-world signed bipartite graphs to verify the effectiveness of our proposed \method.
        For link sign prediction, our method significantly outperforms the existing GNN-based methods for signed bipartite graphs. 
        Furthermore, our method provides a significant advantage over its competitors in terms of computational efficiency, while those methods fail to learn large graphs.
    }
\end{itemize}

The rest of the paper is organized as follows. 
In Section~\ref{sec:related}, we provide a review of previous methods. 
We introduce preliminaries of this work in Section~\ref{sec:preliminaires}. 
After describing our proposed \method in Section~\ref{sec:proposed}, we present our experimental results in  Section~\ref{sec:experiments}.
Lastly, we conclude in Section~\ref{sec:conclusion}.

\section{Related Work}
\label{sec:related}
In this section, we review previous studies for node representation learning in signed graphs.

\subsection{Representation Learning in Signed Unipartite Graphs}
This aims to encode each node into representation (or embedding) vectors by considering positive and negative edges. 
Various encoders have been proposed with the aim of making the embeddings of positively connected nodes similar, while those of negatively connected nodes are dissimilar. 
These encoders are categorized into signed network embedding~\cite{kimWWW2018,xu2022dual,xu2019link} and signed graph neural networks (SGNNs)~\cite{derrICDM2018,liAAAI2020,jungCoRR2020,Huang2021-wp,Ko2023-tv,KoJ24}.
In particular, SGNNs have gained significant attention for their ability to jointly learn downstream tasks in an end-to-end manner.
The early SGNNs were designed for signed unipartite graphs, where all nodes are of the same type. 
As opposed to unsigned graphs, most methods rely on structural balance theory~\cite{cartwright1956structural} to consider signed relationships when designing their own aggregators or additional losses. 
This theory explains how individuals form stable or unstable groups based on their signed relationships, with balanced triads being stable and unbalanced triads leading to tension or change.
For example, SGCN~\cite{derrICDM2018} designs a sign-aware aggregator that aggregates embeddings from potential friends and enemies identified by the balance theory. 
SNEA~\cite{liAAAI2020} exploits a self-attention mechanism to weight each signed edge, thereby improving the sign-aware aggregator following the balance theory.
SGCL~\cite{shuCIKM2021} adopts graph contrastive learning~\cite{YouCSCWS20} with augmented signed graphs following the balance theory. 
However, these methods have limited expressiveness in learning signed bipartite graphs because they do not  consider the unique structure of signed bipartite graphs.


\subsection{Representation Learning in Signed Bipartite Graphs}

More recently, researchers have focused on signed bipartite graphs having two distinct node sets (e.g., users and items). 
To analyze the balance of a signed bipartite graph, Derr et al.~\cite{DerrJCT19} enumerated signed butterfly patterns (i.e., cycles of length 4 between two node types) and found that balanced butterflies appear more frequently than unbalanced ones.
Huang et al.~\cite{huangCIKM2021} developed a sign construction phase that inserts signed edges between nodes of the same type, with the sign determined by the agreement or disagreement between two same-type nodes for a node of a different type.
They then analyzed the balance of the signed graph between either same-type or different-type nodes.
Inspired by this, they proposed SBGNN, which first performs the sign construction phase, then alternates message passing between nodes of the same or those of different types.
Zhang et al.~\cite{zhangSIGIR2023} proposed SBGCL, which introduces graph contrastive learning~\cite{YouCSCWS20} for signed bipartite graphs. 
Similar to SBGNN, SBGCL first performs the sign construction phase on the signed bipartite graph.
The method then randomly perturbs the graph for same-type or different-type nodes as graph augmentations and applies the contrastive learning to the resulting node embeddings from the augmented graphs.
However, these methods significantly increase the number of edges after the sign construction phase, which degrades their efficiency in both time and space.
Furthermore, they simply adopt traditional message passing, which is vulnerable to over-smoothing and noisy interaction issues, thereby limiting their expressive power.

\section{Preliminaries}
\label{sec:preliminaires}
We describe the preliminaries of this work, including notations and problem definition. 
We summarize frequently used symbols in Table~\ref{tab:symbols}.

\setlength{\tabcolsep}{3.3pt}
 \begin{table}
    \small
    \centering
    \caption{
        \label{tab:symbols}
        Frequently-used symbols.
    }
    \begin{tabular}{c|l}
        \toprule
        \textbf{Symbol} & \textbf{Description} \\
        \midrule
        $\graph = (\users, \items, \edges)$ & signed bipartite graph\\
        $\users$ and $\items$ & two sets of nodes of different types such that $\users \cap \items = \emptyset$\\
        $\set{S}$ & set of signs, i.e., $\set{S} \coloneq \{+, -\}$\\
        $\edges$ & set of signed edges, i.e., $\edges \coloneq \{ (u, v, s) | u \in \users, v \in \items, s \in \set{S}\}$ \\
        $d$ & number of features (or embedding dimension)\\
        $\mat{X}_{\users} \in \mathbb{R}^{|\users| \times d}$ & initial feature matrix of nodes in $\users$ \\
        $\mat{X}_{\items} \in \mathbb{R}^{|\items| \times d}$ & initial feature matrix of nodes in $\items$ \\
        $\mat{Z}_{\users} \in \mathbb{R}^{|\users| \times d}$ & final node representations in $\users$ \\
        $\mat{Z}_{\items} \in \mathbb{R}^{|\items| \times d}$ & final node representations in $\items$ \\
        $\matR{s}\in \{0, 1\}^{|\users| \times |\items|}$ & signed biadj. matrix having links of sign $s$ from $\users$ to $\items$ \\
        $\matQ{s}\in \{0, 1\}^{|\items| \times |\users|}$ & signed biadj. matrix having links of sign $s$ from $\items$ to $\users$ \\
        $r$ & ratio of target rank $k$, i.e., $k \defeq \min(|\users|, |\items|) \cdot r$ \\
        $c$ & injection ratio of personalized features \\
        $L$ & number of layers \\
        \bottomrule
    \end{tabular}
\end{table}

\subsection{Notations}
\smallsection{Vector and matrix}
We use lowercase boldface letters (e.g., $\vect{h}$) for vectors, and uppercase boldface letters (e.g., $\mat{A}$) for matrices.
We denote $\vect{h}(i)$ as the $i$-th entry of the vector $\vect{h}$, and $\mat{A}(i, j)$ as the $(i, j)$-th entry of the matrix $\mat{A}$.
We use $\mat{A}(i)$ to represent the $i$-th row vector of the matrix $\mat{A}$.

\smallsection{Signed bipartite graph}
A signed bipartite graph is denoted by $\graph = (\users, \items, \edges)$, where $\users$ and $\items$ are two disjoint and independent sets (or parts) of nodes.  
The set $\edges$ contains signed edges where every edge connects a node in $\users$ and to one in $\items$, and has a positive ($+$) or negative ($-$) sign.
Let $\mat{X}_{\users} \in \mathbb{R}^{|\users| \times d}$ and $\mat{X}_{\items} \in \mathbb{R}^{|\items| \times d}$ be initial node feature matrices for $\users$ and $\items$, respectively, where $d$ is the number of features, and $\mat{X}_{\users}(u)$ indicates the node feature vector of node $u \in \users$ (similar for $\items$).

\smallsection{Signed biadjacency matrices}
Let $\matR{s}\in \{0, 1\}^{|\users| \times |\items|}$ be a signed biadjacency matrix having links of sign $s$ from $\users$ and $\items$ where $s$ indicates the sign in $\{+, -\}$.
Specifically, if there is a link with sign $s$ from a node $u \in \users$ to a node $v \in \items$, then $\matR{s}(u, v) = 1$. 
If there is no link between them, then $\matR{s}(u, v) = 0$.
In the opposite direction, we denote $\matQ{s} \in \{0, 1\}^{|\items| \times |\users|}$ as another signed biadjacency matrix having signed links from $\items$ to $\users$.
Note that $\matabsR$ and $\matabsQ$ have links regardless of their signs, i.e., 
\begin{equation*}
    \matabsR \defeq \matR{+} + \matR{-} \quad \text{and} \quad  \matabsQ \defeq \matQ{+} + \matQ{-}.
\end{equation*}
We denote the transpose of each signed biadjacency matrix as follows:
\begin{definition}[Transpose of Signed Biadjacency Matrices]
For a signed biadjacency matrix $\matR{s} \in \mathbb{R}^{|\users| \times |\items|}$ of sign $s$, its transpose is denoted as $\matRT{s} \defeq (\matR{s})^{\top} \in \mathbb{R}^{|\items| \times |\users|}$.
Similarly, for another matrix $\matQ{s} \in \mathbb{R}^{|\items| \times |\users|}$, its transpose is represented as $\matQT{s} \defeq (\matQ{s})^{\top} \in \mathbb{R}^{|\users| \times |\items|}$.
\end{definition}

\subsection{Problem Definition}
We introduce the formal definition of the problem addressed by this work as follows:
\begin{problem}[Node Representation Learning in Signed Bipartite Graphs]
\label{prob}
Given a signed bipartite graph $\graph = (\users, \items, \edges)$ and initial feature matrices $\mat{X}_{\users}$ and $\mat{X}_{\items}$, the problem aims to learn a representation vector of each node by considering the signed relationships in $\edges$, i.e., it outputs $\mat{Z}_{\users} \in \mathbb{R}^{|\users| \times d}$ and $\mat{Z}_{\items} \in \mathbb{R}^{|\items| \times d}$, the final embedding matrices of nodes in $\users$ and $\items$, from the signed bipartite graph $\graph$.
\qed
\end{problem}

Note that the resulting embeddings in $\mat{Z}_{\users}$ and $\mat{Z}_{\items}$ are fed into a downstream task such as link sign prediction, i.e., it aims to jointly train a classifier that predicts the sign of a link between $u \in \users$ and $v \in \items$ using $\mat{Z}_{\users}(u)$ and $\mat{Z}_{\items}(v)$.

\subsection{Personalized PageRank in Signed Unipartite Graphs}
\label{sec:pre:pagerank}
Signed Random Walk with Restart (SRWR)~\cite{jungICDM2016,JungJK20} is a variant of Personalized PageRank~\cite{pagerank,ShinJSK15,JungPSK17} in signed unipartite graphs, aiming to compute relevance scores between nodes considering their  relationships with various applications~\cite{LeeJ23,ChunLSJ24}.
The main idea is to assign a sign to the random surfer and flip it when moving along a negative edge, allowing the surfer to recognize nodes based on the balance theory\footnote{
The balance theory suggests that networks naturally tend toward balance, where individuals within the same group are more inclined to hold positive sentiments toward one another, while those across different groups are more likely to have negative connections. This creates balanced triadic patterns such as "my friend’s friend is a friend" or "my enemy’s friend is an enemy."
}.
In addition, the positive surfer restarts from one of source nodes with a restart probability $c$, resulting in higher personalized scores for nodes close to $u$.
Given a signed unipartite graph $\mathcal{G} = (\users, \edges)$, suppose its adjacency matrix of sign $s$ is represented as $\matA{s} \in \mathbb{R}^{|\users| \times |\users|}$, and $\matabsA = \matA{+} + \matA{-}$.
Then, $\mattA{s}$ is the row-wise semi-normalized matrix, i.e., $\mattA{s} \defeq \mat{D}_{\users}^{-1}\matA{s}$~\cite{jungICDM2016} where $\mat{D}_{\users}$ is the diagonal degree matrix of $\mathcal{G}$.
The SRWR score vectors are iteratively computed as follows:
\begin{align}
    \begin{split}
        \label{eq:srwr}
        \vect{p}_{\users}^{(l)} &\leftarrow (1-c)\cdot\Big(
            \mattAT{+}\cdot\vect{p}_{\users}^{(l-1)} + \mattAT{-}\cdot\vect{m}_{\users}^{(l-1)}
        \Big) + c\cdot\vect{x}_{\users}, \\
        \vect{m}_{\users}^{(l)} &\leftarrow 
            \underbrace{
                (1-c)\cdot
                \Big(
                    \mattAT{-}\cdot\vect{p}_{\users}^{(l-1)} + \mattAT{+}\cdot\vect{m}_{\users}^{(l-1)}
                \Big),
            }_{\substack{\text{Signed} \\ \text{propagation}}}
            \!\!
            \underbrace{
            \color{white}
            +c\cdot\Big(\vect{x}_{\users}\Big)
            }_{\substack{\text{Injection of} \\ \text{personalization}}}
    \end{split}
\end{align}
where $\mattAT{s}$ is the transpose of $\mattA{s}$,
$0 < c < 1$ is the restart probability, and
$\vect{x}_{\users} \in \mathbb{R}^{|\users|}$ is called a query vector (i.e., for $q \in \set{S}$, where $\set{S}$ is the set of source nodes, $\vect{x}_{\users}(q) = 1/|\set{S}|$, and all other entries are zero).
Note that $\vect{p}_{\users}^{(l)}$ (or $\vect{m}_{\users}^{(l)}$) represents the vector of probabilities that the positive (or negative) surfer visits each node after $l$ steps.

Equation~\eqref{eq:srwr} is interpreted as \textit{signed personalized propagation} of the scores following structural balance theory~\cite{cartwright1956structural}. 
For instance, the term $\mattAT{+} \cdot \vect{p}_{\users}^{(l-1)}$ propagates the positive scores in $\vect{p}_{\users}^{(l-1)}$ over the positive edges in $\mattAT{+}$, with the resulting scores contributing to the next positive scores.
This precisely models the idea that a friend of my friend is my friend, as derived from the balance theory, to which the other matrix multiplication terms also conform.
Thus, it performs the signed propagation with $1 - c$ while injecting the personalization scores in $\vect{x}_{\users}$ with $c$ for each step.

\begin{figure}[t!]
    \vspace{-7mm}
    \centering
    \includegraphics[width=\linewidth]{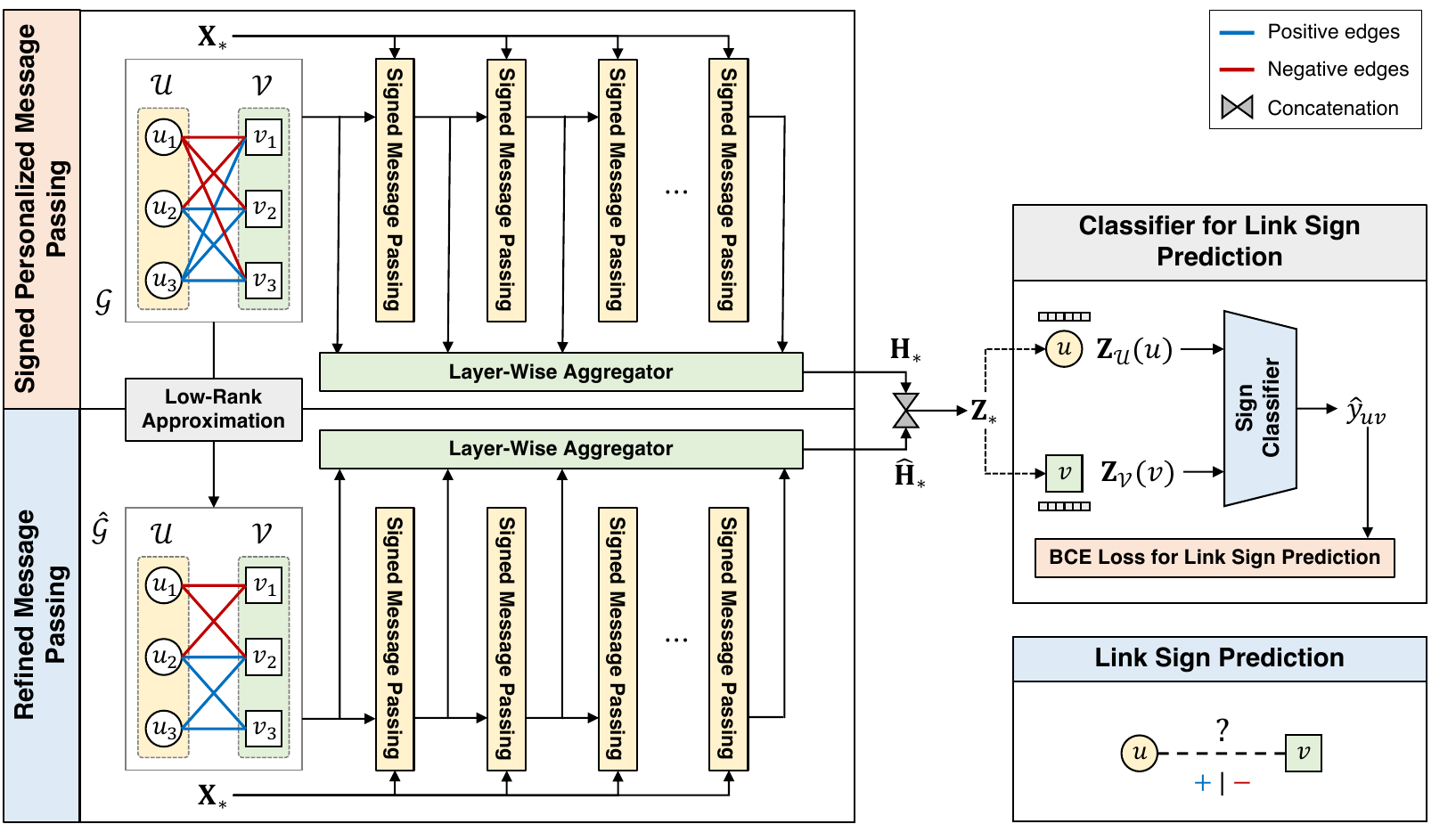}
    \vspace{0.1mm}
    \caption{
    \label{fig:overview}
        Overall architecture of our proposed method \method, which consists of 1) signed personalized message passing and 2) refined message passing on a given signed bipartite graph $\graph$.
        The final node representations in $\mat{Z}_{*}$ are passed to a classifier for downstream tasks, such as link sign prediction.
    }
\end{figure}

\begin{algorithm}[t!]
    \small
    \caption{Our proposed method \method}
    \label{alg:method}
    \begin{algorithmic}[1]
        \Require
        \Statex signed bipartite graph $\graph=(\users, \items, \edges)$
        
        \Statex initial node feature matrices $\XV \in \mathbb{R}^{|\items| \times d}$ and $\XU \in \mathbb{R}^{|\users| \times d}$

        
        \Statex number $L$ of layers 

        \Statex target rank $k$ of a low-rank approximation

        \Statex importance weights $\alpha_{l}$ and $\hat{\alpha}_{l}$ for each layer
        
        \Ensure
        \Statex final node representations $\mat{Z}_{\users}$ and $\mat{Z}_{\items}$

        \vspace{2mm}

        \Statex {\scriptsize $\triangleright$ Preprocessing phase}
        \State set $\Pi_{k} \leftarrow \emptyset$ for storing the preprocessed SVD results with the rank $k$
        \For{\textbf{each} sign $s \in \{+, -\}$}
            \State compute $\matR{s}$ and $\matQ{s}$ from $\graph$
            \State compute row-wise semi-normalized matrices $\mattR{s}$ and $\mattQ{s}$ 
            \State set $\Pi_{k}(\mattRT{s}) \leftarrow \texttt{svd}_{k}(\mattRT{s})$ and $\Pi_{k}(\mattQT{s}) \leftarrow \texttt{svd}_{k}(\mattQT{s})$
        \EndFor
        
        \vspace{2mm}
        \Statex {\scriptsize $\triangleright$ Message passing phase}
        \State compute $\mat{H}_{\users}$ and $\mat{H}_{\items}$ via the signed personalized message passing in Algorithm~\ref{alg:spmp}
        \State compute $\mat{\hat{H}}_{\users}$ and $\mat{\hat{H}}_{\items}$ via the refined message passing in Algorithm~\ref{alg:sloa}

        \State compute $\mat{Z}_{\users} \leftarrow \texttt{concat}(\mat{H}_{\users}, \mat{\hat{H}}_{\users})$ and $\mat{Z}_{\items} \leftarrow \texttt{concat}(\mat{H}_{\items}, \mat{\hat{H}}_{\items})$

        \vspace{2mm} 
        
        \item[] \textbf{return} $\mat{Z}_{\users}$ and $\mat{Z}_{\items}$

    \end{algorithmic}
\end{algorithm}


\section{Proposed Method}
\label{sec:proposed}

In this section, we propose \method, an effective and lightweight method for learning node representations in signed bipartite graphs. 

\subsection{Overview}
\label{sec:proposed:overview}
We summarize the technical challenges for representation learning in signed bipartite graphs as follows:
\begin{enumerate}[noitemsep]
    \item[C1.] {
        \textbf{How to enhance the expressiveness of node representations.}
        As mentioned in Section~\ref{sec:introduction}, the existing methods rely on a naive design for message passing, which is vulnerable to the over-smoothing issue. 
        Additionally, real-world graphs often contain noisy interactions, which hinder the model's ability to effectively learn from the graph.
        How can we address these issues in representation learning on a signed bipartite graph?
    }
    \item[C2.]{
        \textbf{How to efficiently learn node representations.}
        In learning signed bipartite graphs, the previous GNN-based methods add signed edges between nodes of the same type, which can slow down their message passing procedures. 
        Furthermore, these methods include weight matrices of learnable parameters for each layer and additionally optimize their own loss functions, increasing model complexity. 
        How can we design a lightweight method for signed bipartite graphs that avoids unnecessary inefficiencies?
    }
\end{enumerate}

We propose the following ideas to handle the aforementioned challenges:

\begin{enumerate}[noitemsep]
    \item[I1.] {
        \textbf{Signed personalized message passing.}
        We extend the signed personalized propagation of scores to node embeddings in a signed bipartite graph, aiming to mitigate the over-smoothing problem while adhering to balance theory.
    }
    \item[I2.] {
        \textbf{Refined message passing.}
        To reduce the impact of noisy interactions, we apply a low-rank approximation to the signed bipartite graph, creating a refined graph on which we then perform the signed personalized propagation.
        We then integrate the resulting embeddings from the original and refined graphs for better expressiveness.
    }
    \item[I3.] {
        \textbf{Lightweight encoder design.}
        To keep our model lightweight, we avoid adding edges between nodes of the same type, and exclude additional weight matrices at each layer and auxiliary loss optimizations.
    }
\end{enumerate}

We illustrate the architecture of our method \method in Figure~\ref{fig:overview}, and its overall procedure is outlined in Algorithm~\ref{alg:method}.
Given a signed bipartite graph $\graph$ and node feature matrices $\mat{X}_{*}$ for $* \in \{\users, \items\}$, our model \method aims to obtain the final node representations $\mat{Z}_{*}$. 
Our model consists of two encoders: 1) signed personalized message passing and 2) refined message passing. 
The former aims to inject personalized (or input) features during signed message passing, adhering to balance theory (Section~\ref{sec:proposed:spmp}). 
The latter utilizes a low-rank approximation of the graph to perform refined message passing (Section~\ref{sec:proposed:rmp}).

\subsection{Signed Personalized Message Passing on Signed Bipartite Graphs}
\label{sec:proposed:spmp}

\begin{algorithm}[t!]
    \small
    \caption{Singed Personalized Message Passing in \method}
    \label{alg:spmp}
    \begin{algorithmic}[1]
        \Require
        \Statex normalized signed biadjacency matrices $\mattR{s} \in \mathbb{R}^{|\users| \times |\items|}$ and $\mattQ{s} \in \mathbb{R}^{|\items| \times |\users|}$
        \Statex initial node feature matrices $\XV \in \mathbb{R}^{|\items| \times d}$ and $\XU \in \mathbb{R}^{|\users| \times d}$
        \Statex number $L$ of layers
        \Statex importance weight $\hat{\alpha}_{l}$ for each layer
        
        \Ensure
        \Statex hidden embedding matrices $\mat{H}_{\users}$ and $\mat{H}_{\items}$

        \vspace{2mm}
                
        \State initialize $\mat{P}_{*}^{(0)} \leftarrow \mat{M}_{*}^{(0)} \leftarrow \mat{X}_{*}$ for each $* \in \{\users, \items\}$
        
        

        \State set $\mat{P}_{*} \leftarrow \alpha_{0}\cdot\mat{P}_{*}^{(0)}$ and $\mat{M}_{*} \leftarrow \alpha_{0}\cdot\mat{M}_{*}^{(0)}$ for each $* \in \{\users, \items\}$
        
        \For{$l \leftarrow 1$ to $L$}\label{alg:spmp:agg:start}
            \State compute $\PV{l}$ and $\MV{l}$ on $\mattRT{s}$ for sign $s$ \Comment{{\scriptsize Equation~\eqref{eq:spmp:utov}}}
            \State compute $\PU{l}$ and $\MU{l}$ on $\mattQT{s}$ for sign $s$ \Comment{{\scriptsize Equation~\eqref{eq:spmp:vtou}}}
            \State compute $\mat{P}_{*} \leftarrow \mat{P}_{*} + \alpha_{l}\cdot\mat{P}_{*}^{(l)}$ and $\mat{M}_{*} \leftarrow \mat{M}_{*} + \alpha_{l}\cdot\mat{M}_{*}^{(l)}$ for each $* \in \{\users, \items\}$
        \EndFor\label{alg:spmp:agg:end}

        \State compute $\mat{H}_{\users} \leftarrow \texttt{concat}(\mat{P}_{\users}, \mat{M}_{\users})$ and $\mat{H}_{\items} \leftarrow \texttt{concat}(\mat{P}_{\items}, \mat{M}_{\items})$



        
        \vspace{2mm}
        \item[] \textbf{return} $\mat{H}_{\users}$ and $\mat{H}_{\items}$
    \end{algorithmic}
\end{algorithm}

We present the first encoder of our method \method, designed to produce hidden embeddings of nodes for a given signed bipartite graph, with its detailed procedure outlined in Algorithm~\ref{alg:spmp}.
To mitigate over-smoothing and enhance expressiveness, we extend the signed personalized propagation for scores in Equation~\eqref{eq:srwr} to \textit{signed personalized message passing} of embeddings in signed bipartite graphs. 

Suppose we are given the signed bipartite graph $\graph = (\users, \items, \edges)$ and the initial node features $\XU$ and $\XV$.
We aim to obtain hidden embeddings $\mat{H}_{\users}$ and $\mat{H}_{\items}$ of nodes in $\users$ and $\items$, respectively.
%
To account for edge signs, we model two positive and negative embeddings  $\mat{P}_{*}^{(0)}$ and $\mat{M}_{*}^{(0)}$ for each node, initialized with the input node features $\mat{X}_{*}$ where $*$ indicates either of the symbol $\users$ or $\items$. 
Then, we iteratively update them through the signed personalized message passing layers, as described below.

\smallsection{Message passing from $\users$ to $\items$}
Let $\PV{l} \in \mathbb{R}^{|\items| \times d}$ and $\MV{l}  \in \mathbb{R}^{|\items| \times d}$ are matrices of positive and negative embeddings of nodes in $\items$ after $l$ steps (or layers).
Then, they are iteratively computed as follows:
\begin{align}
    \begin{split}
        \label{eq:spmp:utov}
        \PV{l} &\leftarrow 
        (1-c)\cdot\Big(
            \mattRT{+}\cdot\PU{l-1} + \mattRT{-}\cdot\MU{l-1}
        \Big)
        +c\cdot\XV,\\
        \MV{l} &\leftarrow 
        \underbrace{
        (1-c)\cdot\Big(
            \mattRT{-}\cdot\PU{l-1} + \mattRT{+}\cdot\MU{l-1}
        \Big)
        }_{\substack{\text{Signed message passing} \\ \text{based on the balance theory}}}
        \!\!\!\!\!\!
            \underbrace{
            \color{white}
            +c\cdot\Big(\vect{x}_{\users}\Big)
            }_{\substack{\text{Injection of} \\ \text{personalized features}}}
    \end{split}
\end{align}
where $0 < c < 1$ is called the injection ratio of personalized features.
Note that $\mattR{s} \defeq \mat{D}_{\users}^{-1}\matR{s}$ is the row-wise semi-normalized matrix of $\matR{s}$, and its transpose is denoted as $\mattRT{s}$  where $\mat{D}_{\users} = \texttt{diag}(\matabsR\cdot\vect{1})$ is the diagonal degree matrix of nodes in $\users$.


In Equation~\eqref{eq:spmp:utov}, each matrix multiplication is interpreted as a signed message-passing operation based on the balance theory, similar to the approach in Equation~\eqref{eq:srwr}.
For instance, $\mattRT{+}\cdot\PU{l-1}$ propagates the positive embeddings in $\PU{l-1}$ across the positive edges from $\users$ to $\items$ in $\mattRT{+}$, aggregating them as the next positive embeddings for $\items$.
At each step, it injects $\XV$ with a factor of $c$ as local (or personalized) features into the positive embeddings $\PV{l}$, preventing the resulting embeddings from over-smoothing through the iterations.
Note that we inject $\XV$ only into the positive embeddings $\PV{l}$ to distinguish them from the negative embeddings $\MV{l}$. 

\smallsection{Message passing from $\items$ to $\users$}
The positive and negative embeddings of nodes in $\users$ are obtained similarly to Equation~\eqref{eq:spmp:utov}.
Suppose $\PU{l}$ and $\MU{l}$ are the positive and negative embedding matrices of $\users$ at step $l$, respectively.
Then, they are computed as follows:
\begin{align}
    \begin{split}
        \label{eq:spmp:vtou}
        \PU{l} &\leftarrow 
        (1-c)\cdot\Big(
            \mattQT{+}\cdot\PV{l-1} + \mattQT{-}\cdot\MV{l-1}
        \Big)
        +c\cdot\XU,\\
        \MU{l} &\leftarrow 
        (1-c)\cdot\Big(
            \mattQT{-}\cdot\PV{l-1} + \mattQT{+}\cdot\MV{l-1}
        \Big),
    \end{split}
\end{align}
where $\mattQT{s}$ is the transpose of $\mattQ{s} \defeq \mat{D}_{\items}^{-1}\matQ{s}$ and $\mat{D}_{\items}$ is the diagonal degree matrix of nodes in $\items$, i.e., $\mat{D}_{\items} = \texttt{diag}(\matabsQ\cdot\vect{1})$.
Similarly to Equation~\eqref{eq:spmp:vtou}, the local feature $\XU$ is incorporated into the message passing.

\smallsection{Layer-wise aggregation}
In this step, we aggregate the intermediate embeddings from each layer to produce the final output of the signed personalized message passing.
Inspired by LightGCN~\cite{0001DWLZ020}, we integrate the intermediate embeddings using  \textit{layer-wise aggregation} as follows:
\begin{equation}
    \label{eq:spmp:final}
    \begin{aligned}
        \mat{P}_{*} \leftarrow \sum_{l=0}^{L}\alpha_{l}\cdot\mat{P}_{*}^{(l)}
        \quad\text{and}\quad
        \mat{M}_{*} \leftarrow \sum_{l=0}^{L}\alpha_{l}\cdot\mat{M}_{*}^{(l)}, 
    \end{aligned}
\end{equation}
where $*$ represents the symbol $\users$ or $\items$, and $\alpha_{l}$ is the non-negative weight for the $l$-th embedding matrix.
This approach helps further alleviate over-smoothing in the resulting embeddings, and captures the distinct semantics associated with different propagation hops (or layers).
The weight $\alpha_{l}$ can be adjusted as either a hyperparameter or a model parameter to be trained.
Similar to LightGCN, we also set $\alpha_{l}$ uniformly to $1/(L+1)$ to make our method lightweight, and found that this generally provided favorable performance in our preliminary experiments.

After obtaining the positive and negative embeddings, our encoder combines them into the output matrices of hidden embeddings as follows:
\begin{equation}
    \begin{aligned}
        \mat{H}_{\users} \leftarrow \texttt{concat}(\mat{P}_{\users}, \mat{M}_{\users}) \quad\text{and}\quad \mat{H}_{\items} \leftarrow \texttt{concat}(\mat{P}_{\items}, \mat{M}_{\items}), \label{eq:spmp:combine}
    \end{aligned}
\end{equation}
where $\texttt{concat}(\cdot)$ horizontally concatenates its input matrices.

\subsection{Signed Personalized Message Passing on Refined Signed Bipartite Graphs}
\label{sec:proposed:rmp}

\begin{algorithm}[t!]
    \small
    \caption{Refined Message Passing in \method}
    \label{alg:sloa}
    \begin{algorithmic}[1]
        \Require
        \Statex preprocessed SVD results $\Pi_{k}$ for $\mattRT{s}$ and $\mattQT{s}$ with target rank $k$
        
        \Statex initial node feature matrices $\XV \in \mathbb{R}^{|\items| \times d}$ and $\XU \in \mathbb{R}^{|\users| \times d}$
        
        \Statex number $L$ of layers 

        \Statex importance weight $\hat{\alpha}_{l}$ for each layer
        
        
        \Ensure
        \Statex hidden embedding matrices $\mat{\hat{H}}_{\users}$ and $\mat{\hat{H}}_{\items}$ \label{alg:sloa:feature}
        
        \vspace{2mm}
        
        
        
        \State initialize $\hat{\mat{P}}_{*}^{(0)} \leftarrow \hat{\mat{M}}_{*}^{(0)} \leftarrow {\mat{X}}_{*}$ for each $* \in \{\users, \items\}$ \label{alg:sloa:init}
        \State set $\mat{\hat{P}}_{*} \leftarrow \hat{\alpha}_{0}\cdot\mat{\hat{P}}_{*}^{(0)}$ and $\mat{\hat{M}}_{*} \leftarrow \hat{\alpha}_{0}\cdot\mat{\hat{M}}_{*}^{(0)}$ for each $* \in \{\users, \items\}$
        
        \For{$l \leftarrow 1$ to $L$}\label{alg:sloa:agg:start}
        
            \State compute $\hPV{l}$ and $\hMV{l}$ on $\Pi_{k}(\mattRT{s})$ using $\lram$ for sign $s$ \Comment{{\scriptsize Equation~\eqref{eq:rmp:utov}}}
            %
            
            \State compute $\hPU{l}$ and $\hMU{l}$ on $\Pi_{k}(\mattQT{s})$ using $\lram$ for sign $s$ \Comment{{\scriptsize Equation~\eqref{eq:rmp:vtou}}}
            
            \State compute $\hat{\mat{P}}_{*} \leftarrow \hat{\mat{P}}_{*} + \hat{\alpha}_{l}\cdot\hat{\mat{P}}_{*}^{(l)}$ and $\hat{\mat{M}}_{*} \leftarrow \hat{\mat{M}}_{*} + \hat{\alpha}_{l}\cdot\hat{\mat{M}}_{*}^{(l)}$ for each $* \in \{\users, \items\}$
        \EndFor\label{alg:sloa:agg:end}

        \State compute $\mat{\hat{H}}_{\users} \leftarrow \texttt{concat}(\mat{\hat{P}}_{\users}, \mat{\hat{M}}_{\users})$ and $\mat{\hat{H}}_{\items} \leftarrow \texttt{concat}(\mat{\hat{P}}_{\items}, \mat{\hat{M}}_{\items})$

        
        \vspace{2mm}
        
        \item[] \textbf{return} $\mat{\hat{H}}_{\users}$ and $\mat{\hat{H}}_{\items}$ \label{alg:sloa:return}
        
        \item[]\Function{\textnormal{$\texttt{RMP}_{k}$}$\big(\mat{A},  \mat{X}\big)$}{}\label{alg:sloa:lram:start}
        
            \State get the preprocessed SVD result of $\mat{A}$ with the  rank $k$, i.e., $\mat{U}, \mat{\Sigma}, \mat{V} \leftarrow \Pi_{k}(\mat{A})$ 
            \State compute $\mat{T} \leftarrow \mat{U} \big(\mat{\Sigma} (\mat{V}^{\top} \mat{X})\big)$ \Comment{{\scriptsize refined message passing}}
            \State \textbf{return} $\mat{T}$
        \EndFunction\label{alg:sloa:lram:end}
    \end{algorithmic}
\end{algorithm}

We describe another encoder of \method that aims to learn node representations on a refined graph, which is outlined in Algorithm~\ref{alg:sloa}.
As mentioned in Section~\ref{sec:introduction}, real-world bipartite graphs naturally contain noisy interactions, which can potentially harm effective graph learning.
To address this issue, we refine the given signed bipartite graph using low-rank approximation, effectively eliminating such noisy interactions while effectively preserving global structures~\cite{savas2011clustered,caiICLR2023,JungS20}.
Afterward, we apply the signed personalized message passing on the refined signed bipartite graph. 

\smallsection{Preprocessing for low-rank approximation}
We first preprocess the signed bipartite graph to compute its low-rank approximation results, which are calculated once and reused during the subsequent message passing phase.
Specifically, we use truncated singular value decomposition (SVD) to perform the low-rank approximation on the normalized signed biadjacency matrices for each sign $s$, as follows:
\begin{equation}
    \label{eq:svd}
    \begin{aligned}
        \mat{U}_{s}, \mat{\Sigma}_{s}, \mat{V}_{s}^{\top} \leftarrow \texttt{svd}_{k}\left(\mattRT{s}\right)
        \quad\text{and}\quad
        \mat{\bar{U}}_{s}, \mat{\bar{\Sigma}}_{s}, \mat{\bar{V}}_{s}^{\top} \leftarrow \texttt{svd}_{k}\left(\mattQT{s}\right),
    \end{aligned}
\end{equation}
where $k < \texttt{min}(|\users|, |\items|)$ is the target rank of the low-rank approximation. 
The reconstruction (or approximation) of $\mattRT{s}$ is $\mat{U}_{s}\mat{\Sigma}_{s} \mat{V}_{s}^{\top}$ where $\mat{U}_{s}\in\mathbb{R}^{|\items| \times k}$ and $\mat{V}_{s}\in\mathbb{R}^{|\users| \times k}$ are matrices of singular vectors, and $\mat{\Sigma}_{s} \in \mathbb{R}^{k \times k}$ is the diagonal matrix of singular values.
Similarly, $\mattQT{s} \approx \mat{\bar{U}}_{s}\mat{\bar{\Sigma}}_{s} \mat{\bar{V}}_{s}^{\top}$ where $\mat{\bar{U}}_{s}\in\mathbb{R}^{|\users| \times k}$, $\mat{\bar{V}}_{s}\in\mathbb{R}^{|\items| \times k}$, and 
$\mat{\bar{\Sigma}}_{s} \in \mathbb{R}^{k \times k}$. 
Note that the reconstructed matrices form the signed biadjacency matrices of a refined signed bipartite graph.
We define $k \defeq \min(|\users|, |\items|) \cdot r$, where $0 < r < 1$ is the ratio of the target rank.

To reuse the results in the message passing phase, we store them in $\Pi_{k}$ (e.g., a hash map). In other words, $\Pi_{k}(\mattRT{s}) \leftarrow \texttt{svd}_{k}(\mattRT{s})$ and $\Pi_{k}(\mattQT{s}) \leftarrow \texttt{svd}_{k}(\mattQT{s})$.
We employ the randomized SVD algorithm~\cite{halko2011finding,JangCJK18} for efficient SVD computation.

\smallsection{Message passing from $\users$ to $\items$}
The next step is to perform the signed personalized message passing on the refined signed bipartite graph.
The main idea is to replace $\mattRT{s}$ with the reconstruction of $\mat{U}_{s}\mat{\Sigma}_{s} \mat{V}_{s}^{\top}$ in Equation~\eqref{eq:spmp:utov}.
However, the explicit reconstruction incurs inefficiency, as it results in a fully dense matrix, while the original matrix $\mattRT{s}$ is sparse.
Instead, we reorder the matrix multiplications. For example, given an arbitrary matrix $\mat{X} \in \mathbb{R}^{|\users| \times d}$ of embeddings, the associated operations of its message passing are reordered as follows:
\begin{equation*}
    \mattRT{s}\cdot\mat{X} \approx \big(\mat{U}_{s}\mat{\Sigma}_{s} \mat{V}_{s}^{\top}\big)\mat{X} \xRightarrow[\text{reorder}]{} 
    \mat{U}_{s} \big(\mat{\Sigma}_{s} (\mat{V}_{s}^{\top} \mat{X})\big),
\end{equation*}
where the latter is more efficient because the former requires $O\big((k+d)|\users||\items|\big)$ time, while the latter takes $O\big(kd(|\users| + |\items|)\big)$ time.
We formalize this operation as the $\lram$ function, called \textit{refined message passing} (RMP) with the target rank $k$, in Algorithm~\ref{alg:sloa}.
Then, the signed personalized message passing from $\users$ to $\items$ on the refined signed bipartite graph is represented as follows:
\begin{align}
        %
        \hPV{l} &\leftarrow 
        (1-c)\cdot\Big(
            \lram\big(\mattRT{+}, \hPU{l-1}\big) + \lram\big(\mattRT{-}, \hMU{l-1}\big)
        \Big)
        +c\cdot\XV, \nonumber\\
        \hMV{l} &\leftarrow 
        (1-c)\cdot\Big(
            \lram\big(\mattRT{-}, \hPU{l-1}\big) + \lram\big(\mattRT{+}, \hMU{l-1}\big) 
        \Big), \label{eq:rmp:utov}
\end{align}
where $\hPV{l} \in \mathbb{R}^{|\items| \times d}$ and $\hMV{l} \in \mathbb{R}^{|\items| \times d}$ are matrices of positive and negative embeddings of nodes in $\items$ after $l$ steps.

\smallsection{Message passing from $\items$ to $\users$}
Similar to Equation~\eqref{eq:rmp:utov}, the message passing from $\items$ to $\users$ on the refined signed bipartite graph is represented as follows:
\begin{align}
        \hPU{l} &\leftarrow 
        (1-c)\cdot\Big(
            \lram\big(\mattQT{+}, \hPV{l-1}\big) + \lram\big(\mattQT{-}, \hMV{l-1}\big)
        \Big)
        +c\cdot\XU,\nonumber\\
        \hMU{l} &\leftarrow 
        (1-c)\cdot\Big(
            \lram\big(\mattQT{-}, \hPV{l-1}) + \lram\big(\mattQT{+}, \hMV{l-1})
        \Big), \label{eq:rmp:vtou}
\end{align}
where $\hPU{l} \in \mathbb{R}^{|\users| \times d}$ and $\hMU{l} \in \mathbb{R}^{|\users| \times d}$ are matrices of positive and negative embeddings of nodes in $\users$ after $l$ steps.
Note that in Equation~\eqref{eq:rmp:vtou}, the $\lram$ function uses the SVD result of $\mattQT{s}$ for each sign $s$.

\smallsection{Layer-wise aggregation}
After repeating Equations~\eqref{eq:rmp:utov} and~\eqref{eq:rmp:vtou} $L$ times, we aggregate the intermediate results from each layer as follows:
\begin{equation}
    \label{eq:rmp:final}
    \begin{aligned}
        \mat{\hat{P}}_{*} \leftarrow \sum_{l=0}^{L}\hat{\alpha}_{l}\cdot\mat{\hat{P}}_{*}^{(l)}
        \quad\text{and}\quad
        \mat{\hat{M}}_{*} \leftarrow \sum_{l=0}^{L}\hat{\alpha}_{l}\cdot\mat{\hat{M}}_{*}^{(l)}, 
    \end{aligned}
\end{equation}
where $*$ indicates the symbol $\users$ or $\items$, and $\hat{\alpha}_{l}$ is the importance of each layer. 
As described in Equation~\eqref{eq:spmp:final}, we uniformly set $\hat{\alpha}_{l}$ to $1/(L+1)$ to simplify and lighten our model.
Similar to Equation~\eqref{eq:spmp:combine}, we combine the above positive and negative embeddings as follows: 
\begin{equation}
    \begin{aligned}
        \mat{\hat{H}}_{\users} \leftarrow \texttt{concat}(\mat{\hat{P}}_{\users}, \mat{\hat{M}}_{\users}) \quad\text{and}\quad \mat{\hat{H}}_{\items} \leftarrow \texttt{concat}(\mat{\hat{P}}_{\items}, \mat{\hat{M}}_{\items}). \label{eq:rmp:combine}
    \end{aligned}
\end{equation}

\subsection{Final Representations and Loss Function for Link Sign Prediction}
\smallsection{Final representations} 
The ultimate goal of \method is to generate a single representation vector for each node while jointly training the two encoders on the original and refined signed bipartite graphs, respectively, to capture different signals from them simultaneously.
For that, we merge the resulting embeddings from Equations~\eqref{eq:spmp:combine}~and~\eqref{eq:rmp:combine} as follows:
\begin{align}
    \mat{Z}_{\users} \leftarrow \texttt{concat}(\mat{H}_{\users}, \mat{\hat{H}}_{\users}) \quad\text{and}\quad \mat{Z}_{\items} \leftarrow \texttt{concat}(\mat{H}_{\items}, \mat{\hat{H}}_{\items}),
    \label{eq:final}
\end{align}
where $\mat{Z}_{\users}$ and $\mat{Z}_{\items}$ are the matrices of final representations of nodes in $\users$ and $\items$, respectively.


\smallsection{Loss function for link sign prediction} 
As a downstream task in signed bipartite graphs, we aim to predict the sign of a given link $(u, v)$, where $u \in \users$ and $v \in \items$. 
The likelihood score $\hat{y}_{uv}$, indicating that the link $(u, v)$ has a positive sign, is computed as follows:
\begin{equation}
    \hat{y}_{uv} \leftarrow \sigma\Big(
        \texttt{MLP}\Big(
            \texttt{concat}\big(
                \mat{Z}_{\users}(u), \mat{Z}_{\items}(v)
            \big)
        \Big)
    \Big),
    \label{eq:mlp}
\end{equation}
where 
$\mat{Z}_{\users}(u)$ and $\mat{Z}_{\items}(v)$ are the representation vectors of nodes $u$ and $v$, respectively, 
$\sigma(\cdot)$ is the sigmoid function, and $\texttt{MLP}(\cdot)$ is a two-layer multi-layer perceptron module.
The objective function for link sign prediction is based on the binary cross-entropy, defined as follows:
\begin{align}
    \label{eq:loss}
    \mathcal{L}_{\texttt{bce}}(\Phi_{T}; \Theta ) := -\frac{1}{|\Phi_{T}|} \sum_{(u,v) \in \Phi_{T}}y_{uv} \cdot \log{\hat{y}_{uv}} + (1-y_{uv})\cdot\log(1-\hat{y}_{uv}),
\end{align}
where $\Phi_{T}$ is the set of training edges, and $\Theta$ is the set of learnable parameters in our \method. 
The ground-truth sign of the link $(u, v)$ is denoted by $y_{uv}$, where $y_{uv} = 1$ indicates a positive sign, and $y_{uv} = 0$ indicates a negative sign.
To find the optimal parameters of \method while avoiding overfitting on the training dataset, we minimize the following loss function through gradient descent:
\begin{align}
    \mathcal{L}_{\texttt{final}}(\Phi_{T}; \Theta ) := \mathcal{L}_{\texttt{bce}}(\Phi_{T}; \Theta) + \lambda_{\texttt{reg}}\cdot\mathcal{L}_{\texttt{reg}}( \Theta ),
\end{align}
where $\mathcal{L}_{\texttt{reg}}( \Theta )$ is the regularization loss (i.e., $L_2$ norm) for the set $\Theta$ of model parameters, and $\lambda_{\texttt{reg}}$ is the regularization hyperparameter, called weight decay.

\subsection{Computational Complexity Analysis}
In this section, we analyze the computational complexity of our model, \method. Throughout this analysis, we denote the number of edges in $\graph$ as $m = |\edges|$, the maximum number of nodes in $\users$ and $\items$ as $n = \texttt{max}(|\users|, |\items|)$, the rank as $k$, the embedding dimension as $d$, and the number of layers as $L$.
\begin{theorem}[Time Complexity]
The time complexity of Algorithm~\ref{alg:method} is $O(m+n)$ where $L$, $k$, and $d$ are fixed constants.
\end{theorem}
\begin{proof*}
We first analyze the preprocessing phase in Algorithm~\ref{alg:method}, where the bottleneck is the SVD computation in Equation~\eqref{eq:svd}. 
The input matrices are sparse, with fewer non-zero entries than $m$, and their size is $|\users| \times |\items|$.
If we use Randomized SVD for sparse matrices, it takes $O(mk + nk^2)$ time for target rank $k$.
Next, we analyze the message passing phase, which consists of Algorithms~\ref{alg:spmp} and~\ref{alg:sloa}.
For Algorithm~\ref{alg:spmp}, Equations~\eqref{eq:spmp:utov}~and~\eqref{eq:spmp:vtou} takes $O((m+n)dL)$ time because there are eight sparse matrix multiplications, each involving a sparse matrix with fewer non-zero entries than $m$, and they repeats $L$ times.
Note that a single call to the $\lram$ function takes $O(nkd)$ time.
For Algorithm~\ref{alg:sloa}, Equations~\eqref{eq:rmp:utov}~and~\eqref{eq:rmp:vtou} takes $O(nkdL)$ time because there are eight calls for the $\lram$ function and they repeats $L$ times.
Putting everything together, the overall time complexity of Algorithm~\ref{alg:method} is $O\big((m+kn)(k+dL)\big)$. 
Therefore, assuming $L$, $k$, and $d$ are fixed constants, the total time complexity of Algorithm~\ref{alg:method} is $O(m +n)$.
\hfill\qedsymbol
\end{proof*}

Since $k$, $d$, and $L$ are hyperparameters, and assuming they are constant, our method \method is linear in the number of edges and nodes, i.e., it takes $O(m + n)$ time. 
Note that signed bipartite graphs are extremely sparse in the real world, i.e., $m = Cn$ for some constant $C$, and \method takes $O(m)$ time in this case.
As mentioned in Section~\ref{sec:introduction}, both SBGNN and SBGCL add new edges between nodes of the same type if they share common neighbors.
Suppose $M$ is the number of added edges. 
Then, their complexity is proportional to $O(m + M)$, where $M$ is significantly large\footnote{Accurately estimating $M$ is difficult in theory because it depends on the structure of the given graph.
The empirical analysis on $M$ is provided in \ref{sec:app:edges}.
}, which degrades the computational efficiency of SBGNN and SBGCL, while our method requires only $O(m)$ time.

\begin{theorem}[Space Complexity]
The space complexity of Algorithm~\ref{alg:method} is $O(m+n)$ where $L$, $k$, and $d$ are fixed constants.
\end{theorem}
\begin{proof*}
The number of non-zeros in $\mattRT{s}$ or $\mattQT{s}$ less than $m$ in Equations~\eqref{eq:spmp:utov}~and~\eqref{eq:spmp:vtou}; and thus, they takes $O(m)$ space if we use a sparse matrix format.
The results of SVD require $O(nk)$ space in Equation~\eqref{eq:svd}. 
The size of all hidden embedding matrices is $O(nd)$. In Equation~\eqref{eq:mlp}, the $\texttt{MLP}$ module requires $O(d^2)$ space for its parameters.
Thus, the overall space complexity is $O(m + (k+d)n + d^2)$. 
Therefore, assuming $L$, $k$, and $d$ are fixed constants, the total space complexity of Algorithm~\ref{alg:method} is $O(m +n)$.
\hfill\qedsymbol
\end{proof*}


 
    


\section{Experiments}
\label{sec:experiments}

We conducted experiments to answer the following questions:
\begin{enumerate}[noitemsep]
    \item[Q1.] {
        \textbf{Predictive performance.} How effective is our proposed method \method for link sign prediction compared to its competitors?
    }
    \item[Q2.] {
        \textbf{Computation efficiency.} How efficient is \method for training a given signed bipartite graph compared to the baselines?
    }
    \item[Q3.] {
        \textbf{Ablation study.} 
        How much does each component of \method contribute to the performance?
    }
    \item[Q4.] {
        \textbf{Effect of hyperparameters.} 
        How do the hyperparameters of \method affect its performance?
    }
\end{enumerate}

\subsection{Experimental Setting}

\def\arraystretch{1.1} 
\setlength{\tabcolsep}{8.2pt}
\begin{table}[t]
\small
\caption{
    Data statistics of signed bipartite graphs.
}
\label{tab:data}
\centering
\begin{tabular}{crrrrr}
\hline
\toprule
\textbf{Dataset}      & $|\users|$ & $|\items|$ & $|\edges|$  & $|\edges_{+}|$ & $|\edges_{-}|$ \\
\midrule
\texttt{Review}    & 182           & 304           & 1,170     & 464 {\scriptsize (40.3\%)}            & 706 {\scriptsize (59.7\%)}          \\
\texttt{Bonanza}   & 7,919         & 1,973         & 36,543    & 35,805 {\scriptsize ({\color{white}0.}98\%)}       & 738 {\scriptsize ({\color{white}0...}2\%)}         \\
\texttt{ML-1M}     & 6,040         & 3,706         & 1,000,209 & 836,478 {\scriptsize (83.6\%)}      & 163,731 {\scriptsize (16.4\%)}     \\
\texttt{Amazon-DM} & 11,796        & 16,565        & 169,781   & 165,777 {\scriptsize (97.6\%)}      & 4,004 {\scriptsize ({\color{white}0}2.4\%)} \\
   \bottomrule
    \hline
\end{tabular}
\end{table}

\smallsection{Datasets}
For our experiments, we use four real-world signed bipartite graphs, whose statistics are summarized in Table~\ref{tab:data}.
The \texttt{Review} dataset~\cite{HuangSHC21} contains peer review decisions from a computer science conference, where nodes represent reviewers and papers, and edges indicate whether a review was positive (at or above \textit{weak accept}) or negative (at or below \textit{weak reject}).
The \texttt{Bonanza} dataset~\cite{bonanza} has user feedback to sellers on Bonanza, an e-commerce platform, where users and sellers are nodes, and positive or negative ratings given by users to sellers are represented as signed edges.
The \texttt{ML-1M} dataset~\cite{HarperK16} is derived from Movielens and includes user-movie ratings, where users and movies are nodes.
The \texttt{Amazon-DM} dataset~\cite{HeM16, McAuleyTSH15} contains positive or negative user reviews of digital music.
Following previous studies~\cite{huangCIKM2021,zhangSIGIR2023}, ratings of 3 or higher are treated as positive edges, while ratings below 3 are considered negative edges in both the \texttt{ML-1M} and \texttt{Amazon-DM} datasets.

\smallsection{Competitors}
Since \method is specialized for signed bipartite graphs, we compare it with graph neural networks designed for bipartite graphs.
To verify the effectiveness of \method, we compare it with \textbf{SBGNN}\cite{huangCIKM2021} and \textbf{SBGCL}\cite{zhangSIGIR2023}, which are state-of-the-art methods specifically designed for signed bipartite graphs. 
We exclude unipartite signed GNNs, as the previous studies \cite{huangCIKM2021,zhangSIGIR2023} have empirically shown that they underperform in learning signed bipartite graphs.
Furthermore, we compare \method with \textbf{LightGCN}~\cite{0001DWLZ020} and \textbf{LightGCL}~\cite{caiICLR2023}, which are GNN-based methods for unsigned bipartite graphs, to investigate the importance of using the information of edge sign. 
To do this, we omit the edge signs when training LightGCN and LightGCL.

\smallsection{Training and evaluation protocols}
We jointly train each model with a classifier for link sign prediction whose loss function is represented as Equation~\eqref{eq:loss}. 
We randomly split each dataset into 85\% for training, 5\% for validation, and the remaining 10\% for testing.
For a given configuration of hyperparameters and evaluation metric, we train the model using the training dataset and evaluate its accuracy on the validation set.
We select the hyperparameters with the best validation performance and use the corresponding model to measure test accuracy.
Since the target task is binary classification, we use AUC, Binary-F1, Macro-F1, and Micro-F1 scores as evaluation metrics.
Especially, we focus on AUC and Macro-F1 because the datasets exhibit class imbalance as shown in Table~\ref{tab:data}.
We repeat each experiment five times using different random seeds and report the average and standard deviation of the test accuracies.

\smallsection{Hyperparameter tuning}
We set the dimension of the final node representation to 32 and the number of epochs to 200 for all methods to ensure a fair comparison.
We use the Adam optimizer with a learning rate of $5 \cdot 10^{-4}$ and a weight decay of $10^{-5}$.
For the other hyperparameters, we perform a grid search within their specified ranges.
For the target rank ratio $r$, we define its range as $\{0.05, 0.1, 0.2, 0.3, 0.4, 0.5\}$.
For the injection ratio $c$ of personalized features, we set its range to $\{0.01, 0.02, 0.15, 0.45, 0.75, 1.0\}$.
For the number of layers $L$, we set its range to $\{0, \dots, 5\}$ with a step size of 1.
For the hyperparameters of each competitors, we follow their ranges specified in the corresponding paper.



\smallsection{Machine and implementation}
We implement our method \method in Python 3.9 using PyTorch 2.0. 
For the baselines, we utilize their publicly available open-source implementations.
All experiments are conducted on a workstation equipped with an NVIDIA RTX A5000 GPU (24GB VRAM).

\subsection{Predictive Performance (Q1)}

\begin{table*}[t!]
    \centering
    \caption{
        \label{tab:performance}
        The performance of link sign prediction. 
        The bold text indicates the best result, while the underlined text indicates the second-best result.
        "Improv." represents the percentage improvement of \method over the best competitor for each metric.
        Our method \method achieves state-of-the-art performance compared to GNN-based methods for signed bipartite graphs.
    }
    \resizebox{\textwidth}{!}{  
    \begin{tabular}{cccccc}
    \toprule
    \multirow{2}{*}{\textbf{Model}} & \multirow{2}{*}{\textbf{Metric}} &  \multirow{2}{*}{\textbf{Review}} &  \multirow{2}{*}{\textbf{Bonanza}} &  \multirow{2}{*}{\textbf{ML-1M}} & \multirow{2}{*}{\textbf{Amazon-DM}} \\
      \\ 
    \midrule
    \multirow{4}{*}{\textbf{LightGCN}~\cite{0001DWLZ020}} 
    & AUC & 0.4897 $\pm$ 0.0348 & 0.6227 $\pm$ 0.0257 & 0.6903 $\pm$ 0.0031 &  0.5697 $\pm$ 0.0105 \\
    & Binary-F1 & 0.4650 $\pm$ 0.0524 & 0.9876 $\pm$ 0.0016 & \underline{0.9107} $\pm$ 0.0007 & \underline{0.9881} $\pm$ 0.0006 \\
    & Macro-F1 & 0.5162 $\pm$ 0.0343 & 0.5147 $\pm$ 0.0070 & \underline{0.4573} $\pm$ 0.0005 & 0.4940 $\pm$ 0.0003 \\
    & Micro-F1 & 0.4949 $\pm$ 0.0524 & 0.9758 $\pm$ 0.0034 & \underline{0.8360} $\pm$ 0.0011 & 0.9764 $\pm$ 0.0012 \\
    \midrule
    \multirow{4}{*}{\textbf{LightGCL}~\cite{caiICLR2023}} 
    & AUC & 0.5167 $\pm$ 0.0446 & \underline{0.6662} $\pm$ 0.0226 & \underline{0.7275} $\pm$ 0.0026 &  0.6336 $\pm$ 0.0126 \\
    & Binary-F1 & 0.4485 $\pm$ 0.0744 & 0.9892 $\pm$ 0.0015 & 0.9106 $\pm$ 0.0007 & 0.9881 $\pm$ 0.0006 \\
    & Macro-F1 & 0.5269 $\pm$ 0.0455 & 0.4973 $\pm$ 0.0056 & 0.4554 $\pm$ 0.0003 & 0.4955 $\pm$ 0.0010 \\
    & Micro-F1 & 0.5525 $\pm$ 0.0310 & 0.9794 $\pm$ 0.0029 & 0.8359 $\pm$ 0.0011 & 0.9766 $\pm$ 0.0012 \\
    \midrule
    \multirow{4}{*}{\textbf{SBGNN}~\cite{huangCIKM2021}} 
    & AUC & 0.6549 $\pm$ 0.0572 & 0.6534 $\pm$ 0.0332 & - & \underline{0.8790} $\pm$ 0.0169 \\
    & Binary-F1 & 0.5420 $\pm$ 0.0675 & 0.9776 $\pm$ 0.0051 & - & 0.9865 $\pm$ 0.0022 \\
    & Macro-F1 & 0.6271 $\pm$ 0.0435 & \underline{0.5401} $\pm$ 0.0066 & - & \underline{0.7108} $\pm$ 0.0126 \\
    & Micro-F1 & 0.6627 $\pm$ 0.0328 & \underline{0.9798} $\pm$ 0.0035 & - & \underline{0.9795} $\pm$ 0.0011 \\
    \midrule
    \multirow{4}{*}{\textbf{SBGCL}~\cite{zhangSIGIR2023}} 
    & AUC & \underline{0.6960} $\pm$ 0.0525 & 0.6274 $\pm$ 0.0291 & - & - \\
    & Binary-F1 & \underline{0.5460} $\pm$ 0.0702 & \underline{0.9897} $\pm$ 0.0017 & - & - \\
    & Macro-F1 & \underline{0.6374} $\pm$ 0.0336 & 0.5077 $\pm$ 0.0174 & - & - \\
    & Micro-F1 & \underline{0.6644} $\pm$ 0.0148 & \textbf{0.9799} $\pm$ 0.0034 & - & - \\
    \midrule
    \multirow{4}{*}{\textbf{\method}(Ours)} 
    & AUC & \textbf{0.7106} $\pm$ 0.0424 & \textbf{0.6838} $\pm$ 0.0198 & \textbf{0.8269} $\pm$ 0.0016 & \textbf{0.9027} $\pm$ 0.0082 \\
    & Binary-F1 & \textbf{0.5822} $\pm$ 0.0470 & \textbf{0.9900} $\pm$ 0.0016 & \textbf{0.9202} $\pm$ 0.0004 & \textbf{0.9897} $\pm$ 0.0006 \\
    & Macro-F1 & \textbf{0.6690} $\pm$ 0.0238 & \textbf{0.5504} $\pm$ 0.0090 & \textbf{0.6833} $\pm$ 0.0028 & \textbf{0.7180} $\pm$ 0.0135 \\
    & Micro-F1 & \textbf{0.6932} $\pm$ 0.0310 & 0.9795 $\pm$ 0.0035 & \textbf{0.8595} $\pm$ 0.0006 & \textbf{0.9802} $\pm$ 0.9802 \\
    \midrule
    \multirow{4}{*}{\textbf{Improv.}} 
    & AUC & 2.10\% & 2.64\% & 13.66\% & 2.70\%\\
    & Binary-F1 & 6.63\% & 0.03\% & 1.04\% & 0.16\% \\
    & Macro-F1 & 4.96\% & 1.91\% & 49.42\% & 1.01\%\\
    & Micro-F1 & 4.33\% & -0.04\% & 2.81\% & 0.07\%\\
    \bottomrule
    \end{tabular}
    }
    \begin{tablenotes}[flushleft]
    \footnotesize
    {\item[] $-$: it indicates out-of-memory errors during training.}
    \end{tablenotes}
\end{table*}

We evaluate the performance of each model on the link sign prediction task. 
From the results in Table~\ref{tab:performance}, we observe the following:
\begin{itemize}
    \item{
        Our proposed method \method successfully processes the datasets in Table~\ref{tab:data} and achieves the best performance among all tested methods.
    }
    \item{
        Especially, our \method outperforms its competitors in terms of AUC and Macro-F1.  
        It provides an improvement of up to 2.64\% in AUC on the \texttt{Review} dataset and up to 4.96\% in Macro-F1 on the \texttt{Bonanza} dataset.
        This indicates that \method provides more expressive representations and performs well on class-imbalanced datasets.
    }
    \item{
        Note that state-of-the-art models, such as SBGNN and SBGCL, fail to process large datasets. 
        SBGCL ran out of memory on both the \texttt{ML-1M} and \texttt{Amazon-DM} datasets, while SBGNN fails on the \texttt{ML-1M} dataset.
        The main reason is that they add new edges between nodes of the same type if they share common neighbors, significantly increasing the number of edges, especially in dense datasets such as \texttt{ML-1M}.
    }
    \item{
        Compared to LightGCN and LightGCL, our method \method shows superior performance, indicating that exploiting edge signs is crucial for encoding node embeddings, even within bipartite graph neural networks.
    }
\end{itemize}

\subsection{Computational Efficiency (Q2)}
\label{sec:exp:efficiency}

\begin{figure}[t!]
    \centering
    \includegraphics[width=0.4\textwidth]{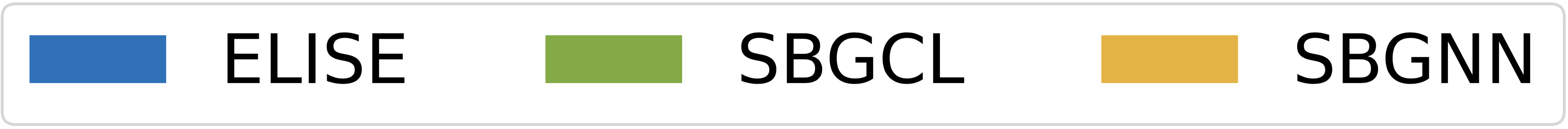}\vspace{0.5mm}\\
    \subfigure[Running time on GPU]{
    \label{fig:comp:gpu}
    \includegraphics[width=0.7\textwidth]{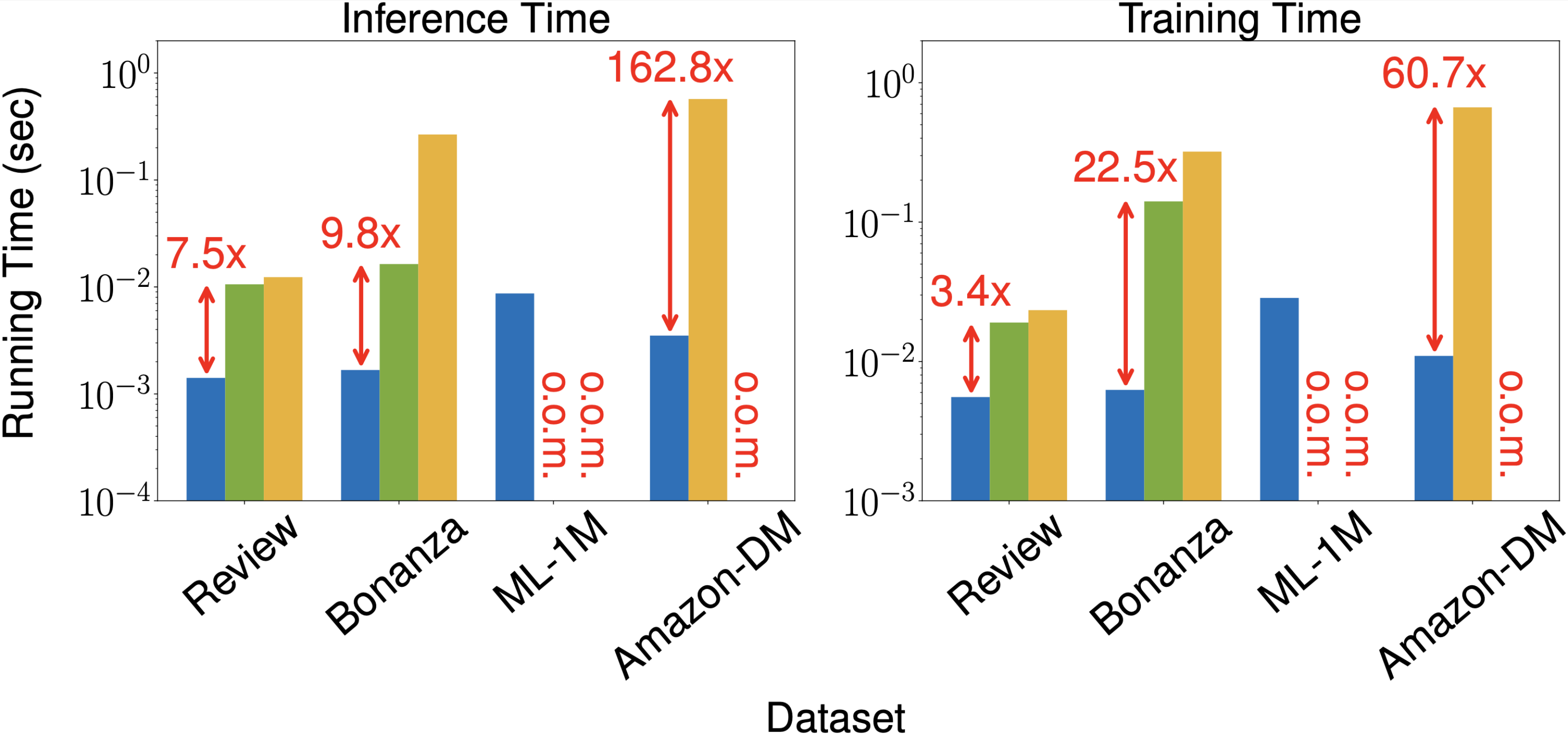}
    }
    \\
    \subfigure[Running time on CPU]{
    \label{fig:comp:cpu}
    \includegraphics[width=0.7\textwidth]{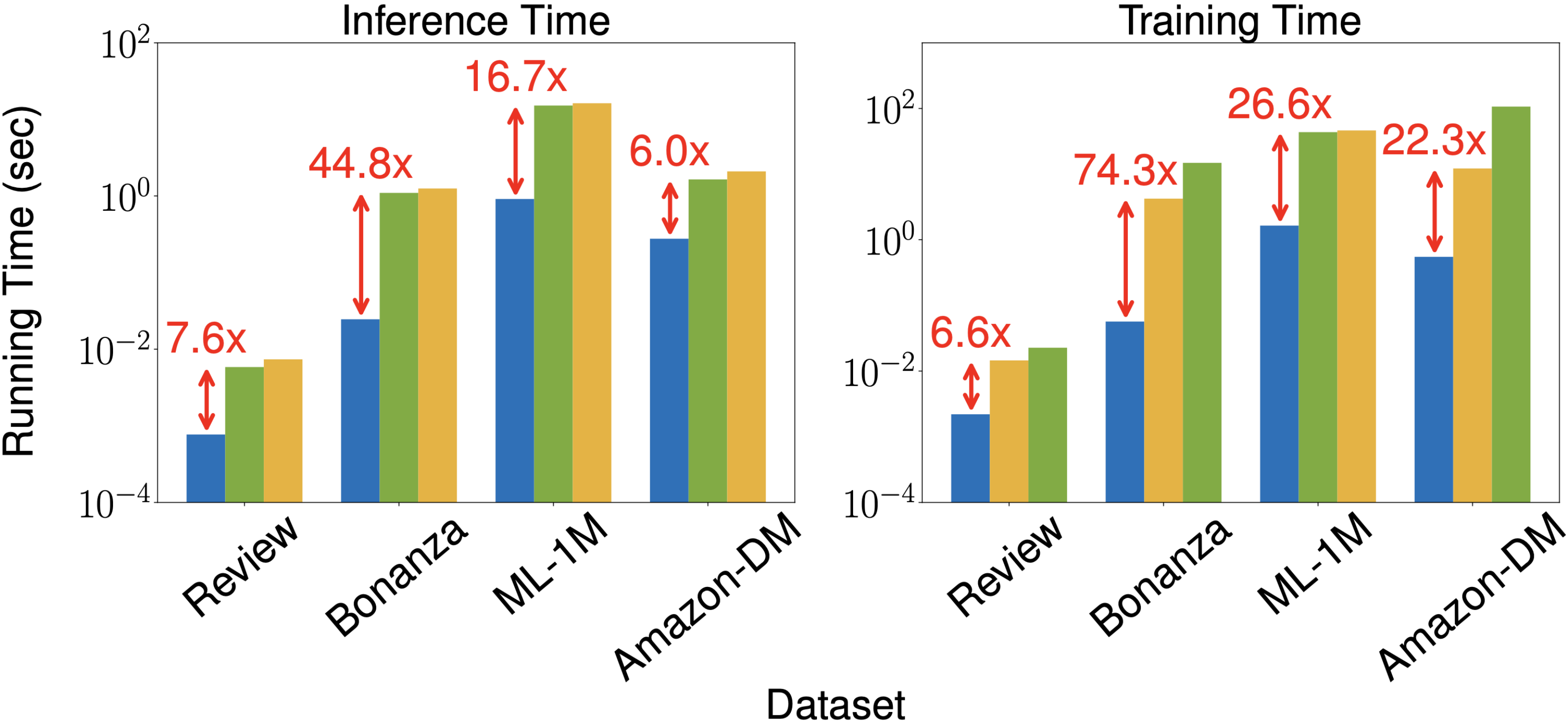}
    }
    \caption{
    \label{fig:time}
    Running time comparison of \method with its competitors, SBGNN and SBGCL, in the inference and training phases, using GPU and CPU, respectively.
    Our method processes all datasets successfully and provides the fastest speed among the tested methods.
    }
\end{figure}

We investigate the computational performance of \method compared to the state-of-the-art models, SBGNN and SBGCL, for signed bipartite graphs.
Specifically, we measure the running time of the inference and training phases for each model on a CPU with a single thread and on a GPU, respectively, to examine performance in each environment (i.e., sequential vs. parallel processing).
We report the average inference and training time over 200 epochs.
Note that the inference time refers to the forward step, while the training time includes both the forward and backward steps, as well as the preprocessing phase (e.g., SVD) for our method.

As shown in Figure~\ref{fig:time}, \method offers the fastest running time in both the inference and training phases across all datasets.
For smaller datasets, such as \texttt{Review} and \texttt{Bonanza}, \method is up 
to $9.8\times$ faster in the inference phase and  $22.5\times$ faster in the training phase than its competitors when using GPU.
Moreover, \method successfully processes larger datasets, such as \texttt{ML-1M} and \texttt{Amazon-DM}, while SBGNN runs out of memory on \texttt{ML-1M} and SBGCL fails on both when using GPU.
The main reason is that both methods add a significantly large number of edges and involve weight matrices for each layer. Furthermore, SBGCL performs graph augmentations, requiring additional memory, which causes it to fail on both datasets.
When using a CPU, SBGNN and SBGCL can also process these datasets, but they show $16.7\times$ slower inference time and $26.6\times$ slower training time than \method.
Note that \method is $162.8\times$ faster in inference and $60.7\times$ faster in training than SBGNN on the \texttt{Amazon-DM} dataset, indicating that our method leverages GPU acceleration more effectively than SBGNN. 




\subsection{Ablation Study (Q3)}

\begin{table*}[t!]
    \centering
    \small
    \caption{
        \label{tab:ablation:val}
        Ablation study of each component of \method in terms of validation performance for link sign prediction where 
        \methodr is a variant of \method without refined message passing, and
        \methods is a variant of \method without signed personalized message passing.
        When jointly learning both components (i.e., \method), it provides the best validation performance across all datasets.
    }
    \begin{tabular}{ccccc}
    \toprule
    \multicolumn{1}{c}{\textbf{Dataset}} & \multicolumn{1}{c}{\textbf{Metric}} & \multicolumn{1}{c}{\textbf{\methodr}} & 
    \multicolumn{1}{c}{\textbf{\methods}} &   \multicolumn{1}{c}{\textbf{\method}} \\
    
    \midrule
    \multirow{2}{*}{\texttt{Review}} 
    & AUC & 0.7187 $\pm$ 0.0679 & 0.7177 $\pm$ 0.0722 & \textbf{0.7390} $\pm$ 0.0774 \\
    & Macro-F1 & 0.6628 $\pm$ 0.0722 & 0.6700 $\pm$ 0.0782 & \textbf{0.6920} $\pm$ 0.0871 \\
    \midrule
    \multirow{2}{*}{\texttt{Bonanza}} 
    & AUC & 0.7269 $\pm$ 0.0314 & 0.6723 $\pm$ 0.0484 & \textbf{0.7460} $\pm$ 0.0450 \\
    & Macro-F1 & 0.5243 $\pm$ 0.0240 & 0.5215 $\pm$ 0.0195 & \textbf{0.5518} $\pm$ 0.0337 \\
    \midrule
    \multirow{2}{*}{\texttt{ML-1M}} 
    & AUC & 0.8148 $\pm$ 0.0039 & 0.8149 $\pm$ 0.0038 & \textbf{0.8259} $\pm$ 0.0017 \\
    & Macro-F1 & 0.6505 $\pm$ 0.0094 & 0.6509 $\pm$ 0.0091 & \textbf{0.6830} $\pm$ 0.0038 \\
    \midrule
    \multirow{2}{*}{\texttt{Amazon-DM}} 
    & AUC & 0.8876 $\pm$ 0.0125 & 0.8812 $\pm$ 0.0143 & \textbf{0.9092} $\pm$ 0.0118 \\
    & Macro-F1 & 0.6950 $\pm$ 0.0266 & 0.6988 $\pm$ 0.0193 & \textbf{0.7295} $\pm$ 0.0120 \\
    \bottomrule
    \end{tabular}%
\end{table*}

\begin{table*}[t!]
    \centering
    \small
    \caption{
        \label{tab:ablation:test}
        Ablation study of each component of \method in terms of test performance for link sign prediction.
    }
    \begin{tabular}{ccccc}
    \toprule
    \multicolumn{1}{c}{\textbf{Dataset}} & \multicolumn{1}{c}{\textbf{Metric}} & \multicolumn{1}{c}{\textbf{\methodr}} & 
    \multicolumn{1}{c}{\textbf{\methods}} &   \multicolumn{1}{c}{\textbf{\method}} \\
    
    \midrule
    \multirow{2}{*}{\texttt{Review}} 
    & AUC & 0.7091 $\pm$ 0.0366 & 0.7032 $\pm$ 0.0387 & \textbf{0.7106} $\pm$ 0.0424 \\
    & Macro-F1 & 0.6628 $\pm$ 0.0256 & 0.6607 $\pm$ 0.0271 & \textbf{0.6690} $\pm$ 0.0238 \\
    \midrule
    \multirow{2}{*}{\texttt{Bonanza}} 
    & AUC & \textbf{0.7235} $\pm$ 0.0257 & 0.6607 $\pm$ 0.0544 & 0.6838 $\pm$ 0.0198 \\
    & Macro-F1 & 0.5260 $\pm$ 0.0157 & 0.5366 $\pm$ 0.0097 & \textbf{0.5504} $\pm$ 0.0090 \\
    \midrule
    \multirow{2}{*}{\texttt{ML-1M}} 
    & AUC & 0.8156 $\pm$ 0.0031 & 0.8157 $\pm$ 0.0030 & \textbf{0.8269} $\pm$ 0.0016 \\
    & Macro-F1 & 0.6515 $\pm$ 0.0058 & 0.6514 $\pm$ 0.0052 & \textbf{0.6833} $\pm$ 0.0028 \\
    \midrule
    \multirow{2}{*}{\texttt{Amazon-DM}} 
    & AUC & 0.8885 $\pm$ 0.0070 & 0.8696 $\pm$ 0.0089 & \textbf{0.9027} $\pm$ 0.0082 \\
    & Macro-F1 & 0.6819 $\pm$ 0.0172 & 0.6838 $\pm$ 0.017 & \textbf{0.7180} $\pm$ 0.0135 \\
    \bottomrule
    \end{tabular}%
\end{table*}

We conduct an ablation study to investigate the effects of the components of \method. 
As described in Section~\ref{sec:proposed}, our method consists of signed personalized message passing and refined message passing.
To evaluate the effectiveness of each component, we consider \methodr and \methods, two variants of \method. 
Specifically, \methodr is a variant without refined message passing, and \methods is a variant without signed personalized message passing.
We report the average AUC and macro-F1 scores for each variant, including \method, across all datasets.
For each variant, we select the best-performing hyperparameters on the validation set and evaluate it on the test set using these hyperparameters.

Tables~\ref{tab:ablation:val} and~\ref{tab:ablation:test} present the results of the ablation study in terms of validation and test accuracy, respectively.
As shown in the tables, \method achieves the best performance in most cases, indicating that jointly learning both components in a single model is beneficial for the evaluation phase, such as validation and testing.
Note that each component provides lower accuracy than \method. 
This indicates that each component has limitations individually, but they exhibit a synergistic effect when signed personalized and refined message passing are jointly considered.
For the \texttt{Bonanza} dataset, \methodr achieves higher test accuracy in AUC than \method, likely due to the gap between validation and test distributions (i.e., the best-performing hyperparameters on the validation set do not yield the best test accuracy).



\subsection{Effect of Hyperparameters (Q4)}
\begin{figure}[t!]
    \centering
    \includegraphics[width=0.6\textwidth]{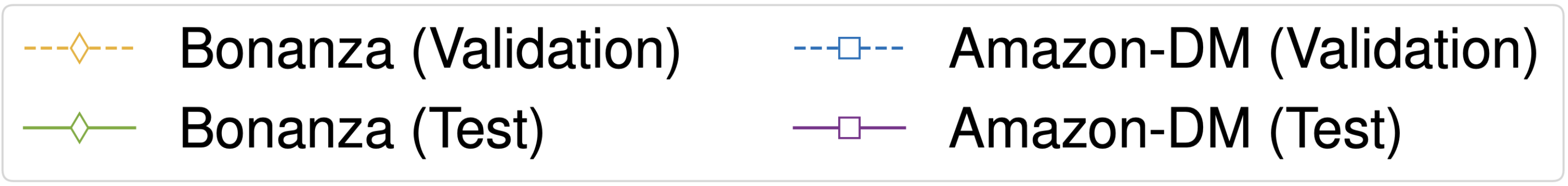}\hspace{-8mm}\vspace{-2mm}\\
    \subfigure[Rank ratio]{
    \hspace{-5mm}
    \includegraphics[width=0.3155\textwidth]{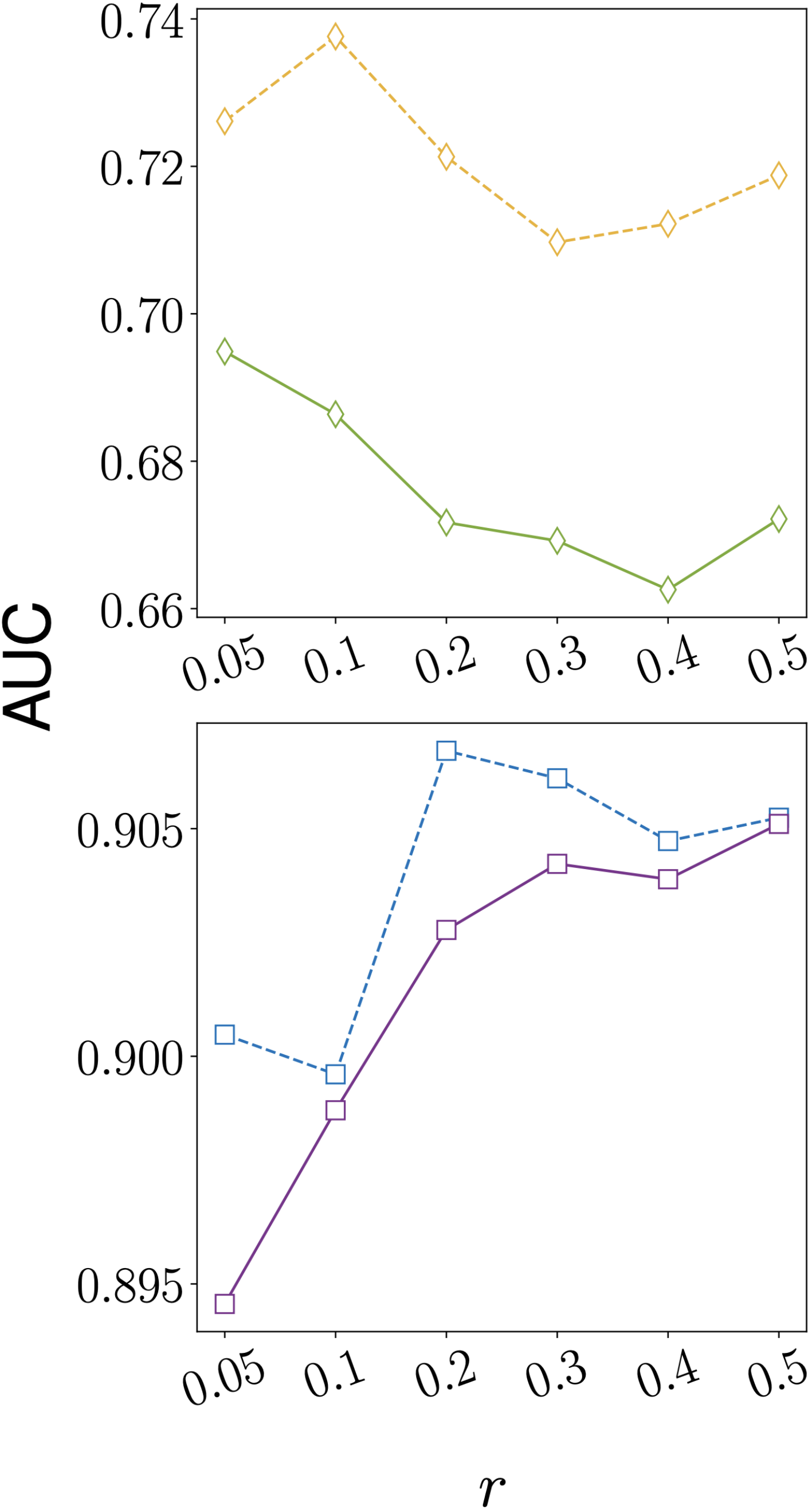}
    \label{fig:param:rank}
    \hspace{-4mm}
    }
    \subfigure[Injection ratio]{
    \hspace{-2mm}
    \includegraphics[width=0.276\textwidth]{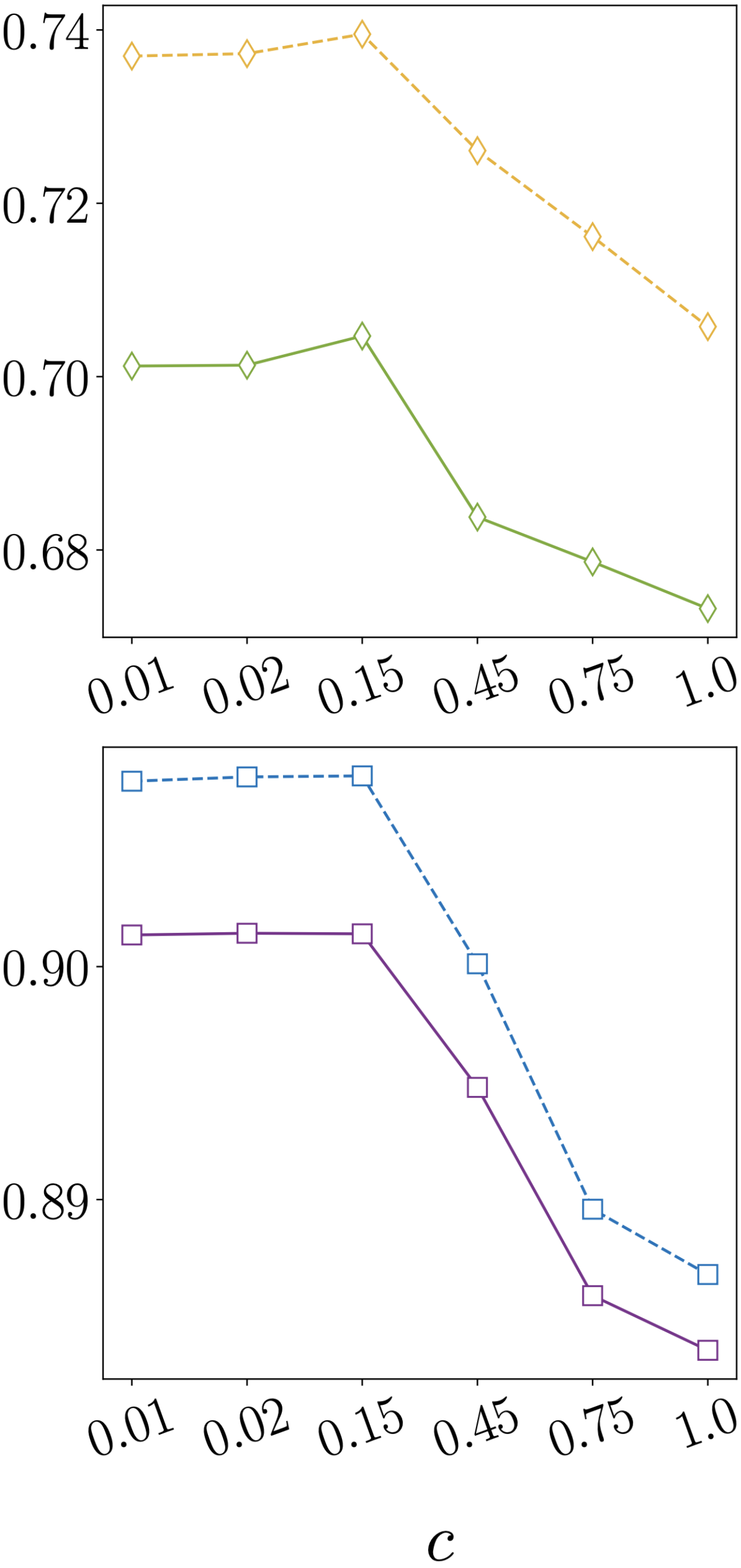}
    \label{fig:param:restart}
    \hspace{-4mm}
    }
    \subfigure[Number of layers]{
    \hspace{-2mm}
    \includegraphics[width=0.283\textwidth]{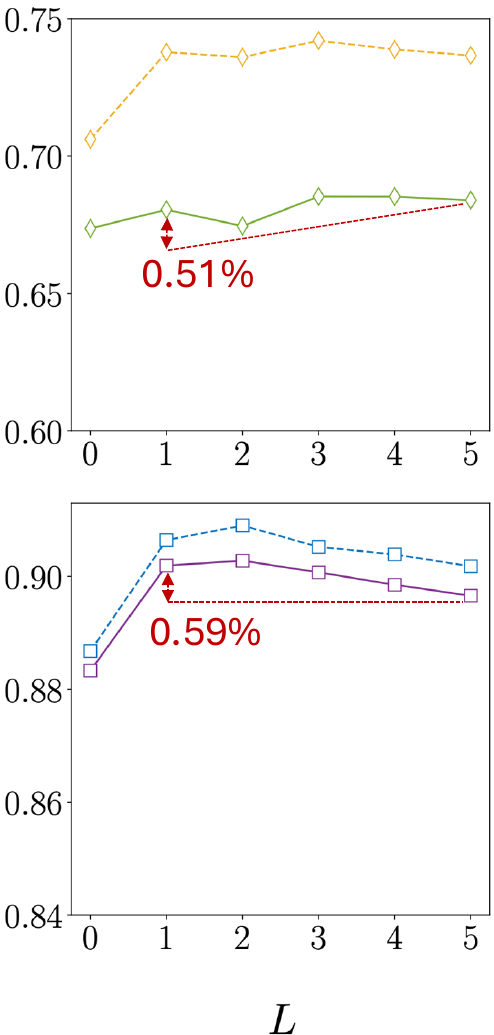}
    \label{fig:param:layer}
    \hspace{-3mm}
    }
    \caption{
    \label{fig:parameter}
    Effect of hyperparameters of \method where (a) $r$ is the rank ratio, (b) $c$ is the injection ratio of personalized features, and (c) $L$ is the number of layers.
    }
\end{figure}

We analyze the effect of hyperparameters of \method. 
As described in Section~\ref{sec:proposed}, \method has three main hyperparameters: the rank ratio $r$, the injection ratio $c$ of personalized features, and the number $L$ of layers.
We vary the range of each hyperparameter, and measure validation and test AUC accuracy of \method on the $\texttt{Bonanza}$ and $\texttt{Amazon-DM}$ datasets.

\smallsection{Effect of the rank ratio} 
Figure~\ref{fig:param:rank} shows the effect of the rank ratio $r$, which controls the target rank of the truncated SVD for low-rank approximation (i.e., $k \defeq \min(|\users|, |\items|) \cdot r$ where $k$ is the target rank).
As shown in the figure, the value of $r$ needs to be properly tuned, as its effect depends on the datasets.
On the \texttt{Bonanza} dataset, both validation and test accuracies decrease as $r$ increases, after peaking at $r = 0.05$ or $r = 0.1$.
In contrast, they tend to gradually increase as $r$ increases on the \texttt{Amazon-DM} dataset, but selecting too large a value of $r$ is undesirable because it results in SVD computations consuming a large amount of memory.
Hence, it is important to carefully select the value of $r$ while considering this trade-off. 

\smallsection{Effect of the injection ratio} 
The effect of the injection ratio $c$, which controls the strength of the injection of personalized features in each message passing, is presented in Figure~\ref{fig:param:rank}.
Note that the performance on both validation and test sets decreases as $c$ increases, peaking at $c = 0.15$ on both datasets. 
This indicates that properly injecting personalized features is helpful for the performance. 
However, it is undesirable to excessively inject them. For example, when $c = 1.0$ (i.e., injecting them without message passing), it results in the worst performance.

\smallsection{Effect of the number of layers}
Figure~\ref{fig:param:layer} shows the effect of the number $L$ of layers, which indicates the number of message passing propagations in \method.
The performance improves initially and then stabilizes, with little significant change as $L$ increases.
This indicates that our signed personalized message passing alleviates the over-smoothing issue in learning signed bipartite graphs, while other methods, such as SBGNN, suffer from this issue, as shown in Figure~\ref{fig:oversmoothing}.

\section{Conclusion}
\label{sec:conclusion}
In this paper, we propose \method (Effective and Lightweight Learning for Signed Bipartite Graphs), a novel representation learning method for signed bipartite graphs. 
Our method consists of two encoders based on signed personalized message passing and refined message passing. 
The former aims to alleviate the over-smoothing issue while satisfying the balance theory, and the latter addresses the noisy interaction issue in learning real-world signed bipartite graphs.
For the first encoder, we extend the signed personalized propagation of scores to node embeddings. 
We further utilize a low-rank approximation to perform refined message passing on the reconstructed graph.
Unlike state-of-the-art methods for signed bipartite graphs, our method is designed to be lightweight because it does not add new edges between nodes of the same type or include weight matrices for each layer.
Through extensive experiments on four real-world signed bipartite graphs, our method achieves higher accuracy in predicting link signs and demonstrates better computational efficiency than its competitors, which fail to process larger datasets.

\section*{Acknowledgements}
{\small
This work was supported by 
National Research Foundation of Korea (NRF) grant by the Korea government (MSIT) (No.2021R1C1C1008526), 
Institute of Information \& communications Technology Planning \& Evaluation (IITP) grant funded by the Korea government (MSIT) (No.RS-2021-II212068, Artificial Intelligence Innovation Hub), and
Cultural Heritage Administration, National Research Institute of Cultural Heritage (No.RS-2024-00398805).
This research was partially supported by the MSIT, Korea, under the Convergence security core talent training business support program (IITP-2024-RS-2024-00426853) and the Innovative Human Resource Development for Local Intellectualization program (IITP-2024-RS-2022-00156360), both supervised by IITP.
}

\appendix

\section{Analysis of the Number of Added Edges}
\label{sec:app:edges}

\begin{table}[h]
\setlength{\tabcolsep}{13pt}
\small
\centering
\caption{
Analysis of how many new edges $B$ are added by the state-of-the-art methods, SBGNN and SBGCL, in comparison to the number $A$ of original edges in the training set.
Note that these methods add a significantly large number of edges.
\label{tab:added_edges}
}
\begin{tabular}{lrrr}
\hline
\toprule
\textbf{Datasets}   & \makecell[r]{$m$: \# of \\ original edges} & \makecell[r]{$M$: \# of \\ added 
 edges} & $M / m$ \\
\midrule
\texttt{Review}    & 994          & 6,890        & 6.9$\times$\\
\texttt{Bonanza}   & 31,061        & 3,426,336     & 110.3$\times$ \\
\texttt{ML-1M}     & 850,177       & 42,736,772    & 50.3$\times$  \\
\texttt{Amazon-DM} & 144,313       & 5,071,652     & 35.1$\times$ \\
\bottomrule
\hline
\end{tabular}
\end{table}

We empirically analyze the state-of-the-art methods, SBGNN and SBGCL, for learning signed bipartite graphs, with a particular focus on the number of new edges added by these methods.
Table~\ref{tab:added_edges} compares the number of original edges with the number of added edges, showing that these methods add a significantly large number of edges. 
Specifically, they add up to $110.3 \times$ more edges than the original on the \texttt{Bonanza} dataset.
This degrades their computational efficiency in both time and space. 
The sparse matrix used to store the edges requires memory proportional to the number of edges.
Also, the time required for message passing is proportional to the number of edges.
Moreover, SBGCL performs graph augmentations for its contrastive learning by copying the graph with added edges and randomly perturbing it. 
This is one of the reasons why these methods fail to process larger datasets when using a GPU, as shown in Section~\ref{sec:exp:efficiency}.

\section{Further Analysis on the Number of Layers}
\label{sec:app:smoothing}
\begin{figure}[h!]
    \centering
    \includegraphics[width=0.45\textwidth]{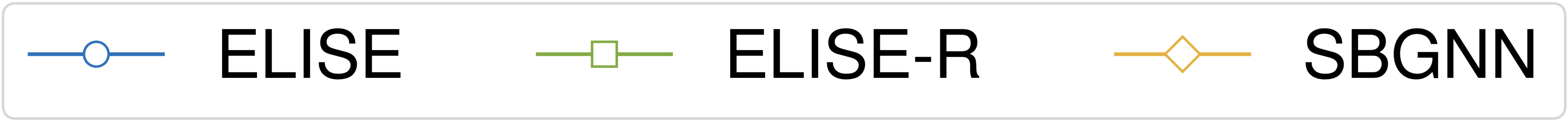}\vspace{0.45mm}\\
    \subfigure{ 
    \label{fig:os:trends}
    \includegraphics[width=0.5\textwidth]{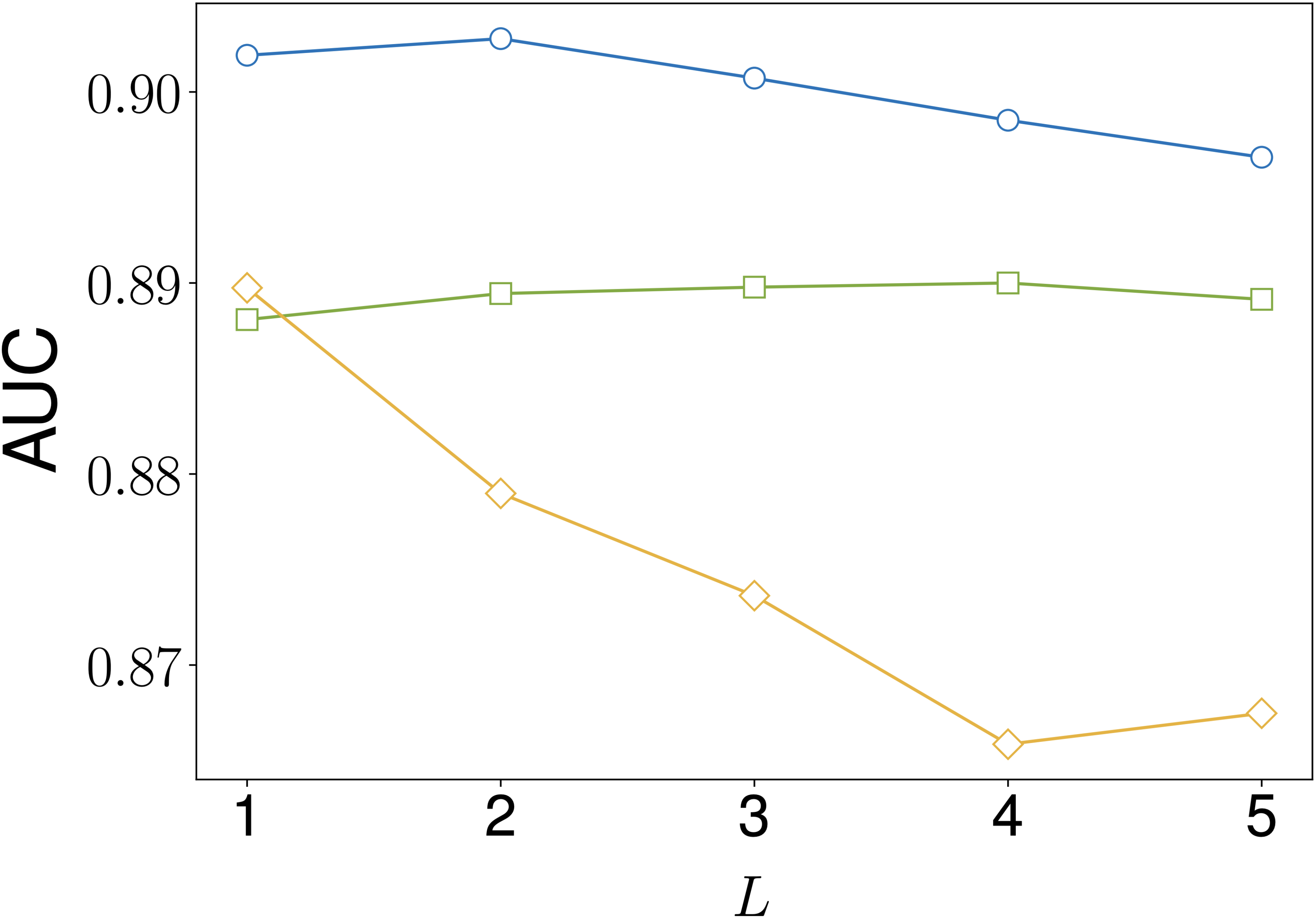}
    }
    \hspace{1cm}
    \caption{
    \label{fig:oversmoothing}
    Effect of the number $L$ of layers on \method, \methodr, and SBGNN for the \texttt{Amazon-DM} dataset, where \methodr is a variant of \method without refined message passing (i.e., it performs only signed personalized message passing).
    Compared to SBGNN, our approaches show more robust performance with respect to $L$, effectively alleviating the over-smoothing issue.
    }
\end{figure}

In this section, we further analyze the effect of the number of layers. 
We compare our method \method and its variant, \methodr, with SBGNN, the state-of-the-art competitor.
\methodr is a variant of \method that excludes refined message passing, performing only signed personalized propagation on the original graph.
We measure test AUC for each model varying the number $L$ of layers from $1$ to $5$ on the \texttt{Amazon-DM} dataset.
As shown in Figure~\ref{fig:oversmoothing}, the performance of SBGNN significantly decreases as $L$ increases, compared to \method and \methodr.
This indicates that SBGNN suffers from the over-smoothing issue during learning.
On the other hand, \methodr shows robust performance with respect to $L$, i.e., the signed personalized message passing effectively alleviates the over-smoothing issue.
Note that \method is relatively more sensitive to $L$ compared to \methodr, even though \method achieves higher accuracy than \methodr.
This suggests that the refined message passing is beneficial for improving performance but is more prone to overfitting due to the increase in model parameters.
Despite this, \method addresses the over-smoothing issue more effectively than SBGNN.

{\small
\bibliographystyle{plain}
\bibliography{myref}
}

\end{document}